\documentclass{article}

\usepackage[preprint]{neurips_2025}

\usepackage{algorithm}
\usepackage[noend]{algpseudocode}

\usepackage[noend]{algpseudocode}

\usepackage[utf8]{inputenc} 
\usepackage[T1]{fontenc}    
\usepackage{hyperref}       
\usepackage{url}            
\usepackage{booktabs}       
\usepackage{amsfonts}       
\usepackage{csquotes}       
\usepackage{natbib} 
\usepackage{textgreek}      
\usepackage{nicefrac}       
\usepackage{microtype}      
\usepackage{xcolor}         

\usepackage{amsmath,amssymb}
\usepackage{graphicx}
\usepackage{wrapfig}
\usepackage{multirow}

\usepackage[most]{tcolorbox}
\usepackage{enumitem}
\tcbset{
  promptbox/.style={
    fonttitle=\bfseries,
    title=Prompt,
    breakable,
    left=1mm, right=1mm, top=1mm, bottom=1mm,
  }
}

\usepackage[most]{tcolorbox}
\usepackage{listings}
\title{From Prompt to Protocol: Fast Charging Batteries with Large Language Models}

\author{
Ge Lei$^{1}$\thanks{g.lei23@imperial.ac.uk} \quad
Ferran Brosa Planella$^{2}$ \quad
Sterling G. Baird$^{3}$ \quad
Samuel J. Cooper$^{1}$\thanks{samuel.cooper@imperial.ac.uk} \\
\small $^{1}$Dyson School of Design Engineering, Imperial College London, London SW7 2AZ, UK \\
\small $^{2}$University of Warwick, Coventry CV4 7AL, UK \\
\small $^{3}$Acceleration Consortium, University of Toronto, 80 St George Street, Toronto, ON M5S 3H6, Canada
}

\begin{document}

\maketitle

\begin{abstract}

Efficiently optimizing battery charging protocols is challenging because each evaluation is slow, costly, and non-differentiable. Many existing approaches address this difficulty by heavily constraining the protocol search space, which limits the diversity of protocols that can be explored, preventing the discovery of higher-performing solutions. We introduce two gradient-free, LLM-driven closed-loop methods: Prompt-to-Optimizer (P2O), which uses an LLM to propose the code for small neural-network–based protocols, which are then trained by an inner-loop, and Prompt-to-Protocol (P2P), which simply write an explicit function for the current and its scalar parameters. Across our case studies, LLM-guided P2O outperforms neural networks designed by Bayesian optimization, evolutionary algorithms, and random search. In a realistic fast-charging scenario, both P2O and P2P yield around a 4.2\% improvement in state of health (capacity retention–based health metric under fast-charging cycling) over a state-of-the-art multi-step constant current (CC) baseline, with P2P achieving this under matched evaluation budgets (same number of protocol evaluations). These results demonstrate that LLMs can expand the space of protocol functional forms, incorporate language-based constraints, and enable efficient optimization in high-cost experimental settings.

\end{abstract}

\section{Introduction}
\label{sec:intro}

Large language models (LLMs) are emerging as powerful tools for tackling problems that are hard to formalize, slow to evaluate, and only partially understood, where classical optimization pipelines struggle to even define a useful search space. In these settings, the design of the solution itself (i.e. the choice of representation, structural priors, and constraints) is often at least as important as numerical parameter tuning, yet is typically encoded through hand-crafted parameterizations or rigid templates. LLMs offer a different route, since they can generate executable algorithms and controllers directly from high-level, language based specifications, incorporate heterogeneous qualitative feedback, and iteratively refine candidate solutions using natural language descriptions of failures and constraints. This reframes difficult scientific design tasks as closed-loop dialogue with a generative model, rather than as pure black-box optimization over a fixed hypothesis class. 

In this work, we instantiate this idea in the context of battery charging protocol design. The optimization of battery charging protocols is fundamental to next-generation electric mobility, because the charging protocol governs how energy is delivered into the cell and directly influences both charging efficiency and long-term battery durability \citep{keil2016charging, mothilal2021impact}. Poorly designed protocols can accelerate degradation or create thermal and mechanical stresses that reduce battery lifespan. This challenge becomes even more pronounced during fast charging: although it shortens downtime, it significantly intensifies degradation mechanisms such as lithium plating \citep{huang2022onboard}, overheating or thermal runaway \citep{cheng2025early, comanescu2024ensuring}. These efficiency-aging trade-offs highlight the need for charging protocols that achieve rapid charging while maintaining battery longevity, a requirement for reliable and high-performance electric mobility.

\begin{figure}[t]
\centering
    \includegraphics[width=1\linewidth]{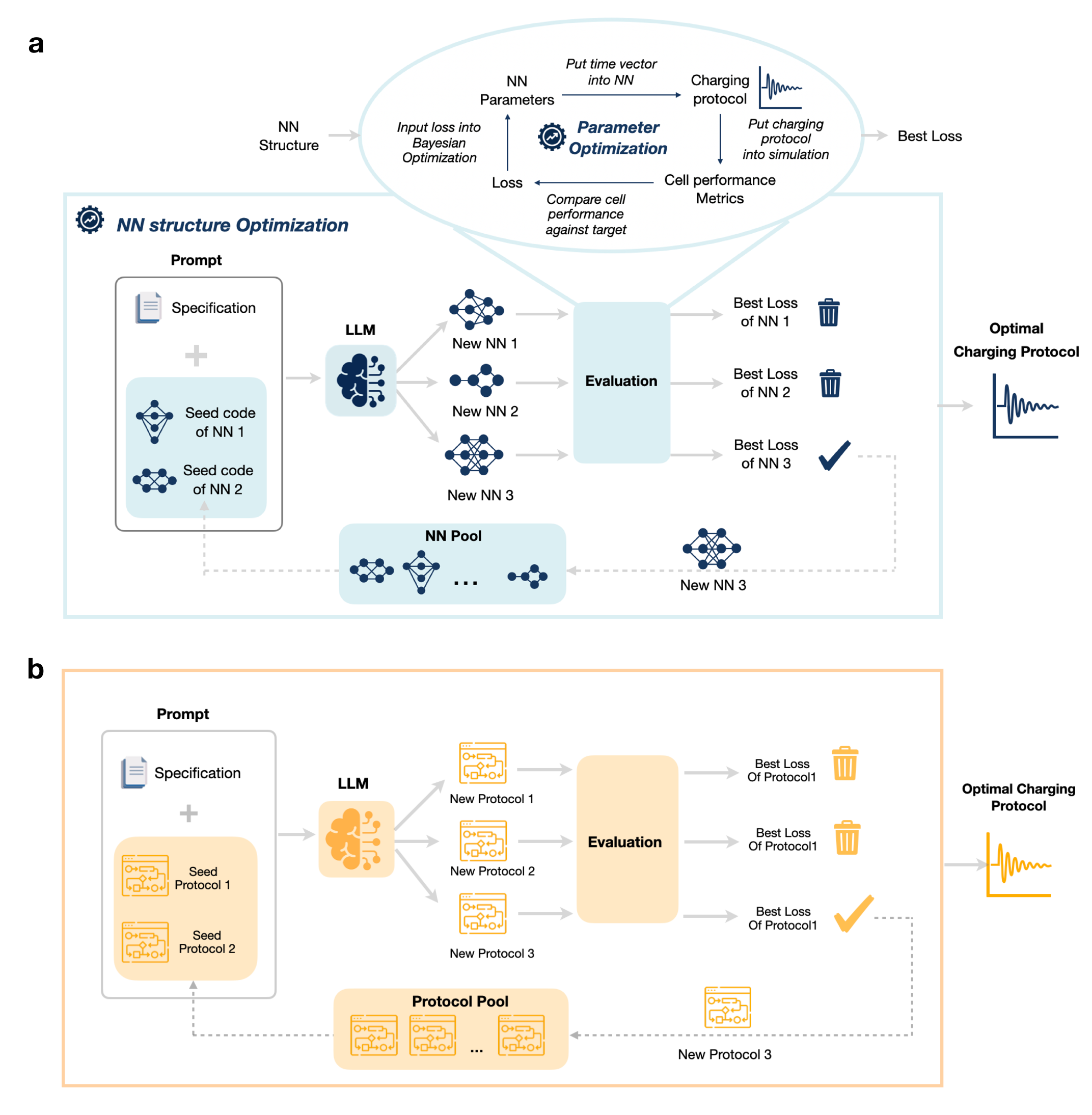}
    \caption{Overview of the proposed LLM-driven optimization frameworks. (a) P2O (two loops): LLM-guided neural network generation and refinement with a SAASBO-based inner optimization loop. (b) P2P (single loop): Direct LLM generation of explicit charging protocols without inner-loop optimization.}
    \label{fig:workflows}
\end{figure} 

However, optimizing battery charging protocols remains a long-standing challenge due to the vast design space and the nonlinear, history-dependent nature of battery dynamics. The difficulty is compounded by the slow, costly, and non-differentiable nature of experimental evaluations, which severely limits how quickly protocols can be tested and improved. Consequently, most existing studies impose strong structural constraints on the protocol class, reducing the search space to make the optimization tractable. Analytical forms such as exponential or sigmoid curves \citep{sikha2003comparison, althurthi2024comparison} are easy to optimize but impose strong inductive biases and limit strategy diversity. Piecewise-constant controls \citep{attia2020closed, kim2024advanced} offer more flexibility, but too few segments oversimplify the space, whereas too many create jagged, unrealistic profiles and lead to high-dimensional non-convex problems. More detailed physics-based approaches have also computed fast-charging protocols directly from high-fidelity porous-electrode models using dynamic optimization or hybrid simulation frameworks (e.g., the MPET-based protocol design of Galuppini \textit{et al.}\cite{galuppini2023efficient}), but these methods still assume fixed operating-mode structures and remain computationally intensive, limiting their flexibility for broad protocol exploration.


LLMs are widely recognized for their language skills \citep{zhao2023survey}, but recent work has shown it has opened up novel approaches to addressing optimization problems (most notably, ``FunSearch") \citep{FUNSEARCH, lange2024large, wu2024evolutionary, wang2025large}. Compared to conventional methods, LLMs offer several notable advantages: (1) they can interpret rich, language-based feedback instead of relying solely on scalar rewards; (2) their strong generative ability allows exploration beyond predefined search spaces; and (3) their broad prior knowledge enables more informed and efficient initialization.

Although several recent studies have demonstrated the ability of LLMs to iteratively improve algorithms, these demonstrations largely rely on settings where algorithm evaluations are extremely fast, enabling the model to perform thousands of trial-and-error iterations. In applications such as battery charging protocol optimization, however, each evaluation is slow, expensive, and often experimentally constrained, raising the question of whether LLM-based optimization strategies can remain effective under such slow-feedback conditions.

At the same time, the intrinsic strengths and limitations of LLMs impose additional considerations. As language-based models, LLMs excel at structural abstraction. Yet they typically exhibit limited sensitivity to numerical information, as evidenced by their well-known difficulties with arithmetic and quantitative precision \citep{qian2022limitations, baeumel2025lookahead, gambardella2024language}. This contrast naturally raises a methodological question: in slow-feedback, high-cost optimization settings where efficiency is paramount, should LLMs primarily focus on designing and refining the algorithmic framework while delegating numerical parameter tuning to dedicated optimizers, or is it more effective to allow LLMs to jointly handle both structural and numerical decisions in an end-to-end manner?

To answer these questions and overcome the limitations of conventional charging protocol parameterizations, we introduce two LLM-driven paradigms: Prompt-to-Optimizer (P2O; two loops) and Prompt-to-Protocol (P2P; single loop), as shown in Fig.\ref{fig:workflows}. Both approaches leverage the reasoning and generative abilities of LLMs to explore broad protocol spaces beyond manually defined search structures. P2O adopts an evolutionary optimization framework in which the LLM iteratively designs and refines neural network–based charging protocols through mutation of parent architectures. While neural networks provide a powerful and flexible representation for exploring broad protocol spaces, their training is hindered by the lack of reliable gradients in real experiments and the instability or computational cost of simulated ones. To address these limitations, P2O employs lightweight neural architectures (typically fewer than 35 parameters) and then optimizes their parameters using a gradient-free Sparse Axis-Aligned Subspace Bayesian Optimization (SAASBO) inner loop \citep{eriksson2021highdimensionalbayesianoptimizationsparse,baird2022high}, which enables exploration of high-dimensional search spaces.

P2P, by contrast, adopts a more direct, end-to-end paradigm: the LLM generates fully executable charging protocols in a single step, specifying both algorithmic structures and numerical parameters without relying on an inner-loop optimizer. While P2O (two loops) emphasizes systematic framework evolution and sample-efficient parameter tuning, P2P enables rapid synthesis of diverse, ready-to-evaluate strategies. Together, these two approaches span a spectrum of autonomy and control, providing a flexible framework for integrating LLMs into battery charging optimization workflows.

To evaluate the effectiveness and generality of the proposed methods, we conduct three representative case studies: predefined heat search, constant-heat protocol search, and degradation-aware fast-charging protocol search. In the outer-loop architecture evolution tasks, LLM-guided P2O consistently outperforms Bayesian optimization, evolutionary algorithms, and random search. In the most challenging fast-charging scenario, both P2P (single loop) and P2O (two loops) achieve strong performance, surpassing a recent state-of-the-art baseline by approximately 4.2\% in final SOH. P2P reaches its best performance using the same number of experimental trials as the baseline. P2O exceeds the baseline under the same trial budget and achieves the best overall performance after further evolutionary refinement. Overall, these results suggest that LLM-based optimization provides a practical alternative to conventional parameterized approaches, even under slow-feedback conditions, and can be integrated into battery-protocol design without requiring large trial budgets.

\section{Methods}

Some existing approaches build charging protocols using bespoke functions selected by domain experts, often making reference to physical phenomena (e.g., exponentials to capture Arrhenius behavior). However, domain expertise is generally only able to handle a modest degree of complexity and dimensionality before intuition fails. For example, it may be desirable to incorporate information from many input parameters, such as temperatures, voltages, or derived metrics such as the state-of-charge. 

Another family of approaches is to optimise a piecewise function comprised of constant current steps. While simple to implement, this approach restrict expressiveness. To overcome these limitations, we propose two complementary approaches to expand the search space and improve optimization efficiency: Prompt-to-Optimizer (P2O; two loops), which employs LLMs to design optimization frameworks that iteratively refine charging protocols through an inner-loop Bayesian optimization; and Prompt-to-Protocol (P2P; single loop), where the LLM directly generates executable charging strategies with both algorithmic structures and parameter values in a single step.

\subsection{Prompt-to-Optimizer (P2O)}

Neural networks are highly expressive, enabling smooth and flexible charging protocols across broad design spaces. This is because their outputs vary continuously with respect to their parameters, so small parameter updates tend to produce gradual, well-behaved changes in the generated protocol. However, training them typically requires reliable gradients. Real battery experiments, unfortunately, do not provide differentiable feedback, which makes gradient-based learning infeasible.

Although simulations are in principle capable of producing gradients, in practice these gradients are typically not suitable for optimization. Electrochemical battery models involve stiff ODE/PDE dynamics \citep{kim2021stiff}, making gradients unstable and highly sensitive to small perturbations. Long-horizon differentiation through implicit solvers demands substantial memory and computation \citep{gholami2019anode}, and the resulting gradients frequently fail to provide useful descent directions in highly nonconvex landscapes \citep{dauphin2014identifying}. In addition, voltage and temperature safety limits induce abrupt mode transitions, and multiple termination conditions (i.e., voltage cut-offs, thermal thresholds) introduce discrete events that are inherently non-differentiable \citep{hussein2011review, kim2024accounting}. These properties are intrinsic to battery operation and remain present even when protocols are smoothed. As a result, gradient-based training of neural networks remains highly challenging for battery charging protocol optimization.

For small neural networks (e.g. 35 parameters), gradient-free optimization methods such as SAASBO can offer an alternative to gradient-based training. Although each small network may have limited expressiveness, evolving a diverse set of small architectures over time enables exploration of a broader solution space while preserving optimization stability.

\begin{algorithm}[ht]
\caption{P2O (two-loop) Evolutionary Optimization}
\label{alg:llm_saasbo}
\begin{algorithmic}[1]
\Require Initial pool $\mathcal{P}_0$ (set of $(s,\theta,L)$ tuples or architectures with initialized losses); 
         generations $T$; children per generation $C$;
         pool size $N$; SAASBO budget $B$
\Ensure Best network structure and parameters
\State $\mathcal{P} \gets \mathcal{P}_0$
\For{$t = 1$ to $T$}
    \State $\mathcal{C} \gets \emptyset$ \Comment{reset child set}
    \For{$c = 1$ to $C$}
        \State Select parents $p_1,p_2$ from $\mathcal{P}$ \Comment{e.g.\ tournament selection}
        \State $s_c \gets \textsc{LLMGenerate}(p_1,p_2)$ 
        \State $(\theta_c^\ast,L_c) \gets \textsc{SAASBOOptimize}(s_c,B)$ 
        \Comment{optimize parameters via SAASBO for $B$ evaluations}
        \State $\mathcal{C} \gets \mathcal{C} \cup \{(s_c,\theta_c^\ast,L_c)\}$
    \EndFor
    \State $\mathcal{P} \gets \textsc{SelectTopN}(\mathcal{P}\cup\mathcal{C},N)$ 
    \Comment{keep $N$ best networks by loss}
\EndFor
\State \Return $(s^\ast,\theta^\ast)$ 
\Comment{where $(s^\ast,\theta^\ast,L^\ast)=\arg\min_{(s,\theta,L)\in\mathcal{P}}L$}
\end{algorithmic}
\end{algorithm}

Specifically, we propose an evolutionary optimization framework to efficiently discover optimal charging protocols consisting of two nested loops: an outer loop that employs LLM-based neural architecture generation, and an inner loop in which the networks parameters are optimized with SAASBO, as illustrated in Fig. \ref{fig:workflows}. The process begins with an LLM generating an initial pool of neural network architectures, based on a prompt that specifies various consideration, including the constraint that every network should contain fewer than 35 trainable parameters. These architectures directly correspond to candidate charging-protocol structures, expressed as Python code. At each subsequent generation, two high-performing networks are sampled from the current pool (e.g., via tournament selection) to serve as ``parents". By prompting the LLM to propose new architectures based on the two selected parents, the model generates new child architectures.

In the inner loop, for each newly generated architecture, its trainable parameters are optimized using SAASBO, a Bayesian optimization method designed to remain effective in high-dimensional search spaces. After optimization, the resulting performance metric (best loss) for each network is compared against those of the architectures currently in the pool. If the new network achieves a better loss than any existing member, it replaces the weakest architecture. Consequently, the best-performing networks are retained for the next generation, while lower-performing ones are gradually replaced. Through this iterative process of generation, optimization, and selection, the performance of the pool of protocols gradually improves. The complete procedure is summarized in Algorithm \ref{alg:llm_saasbo}. The detailed prompt is provided in Appendix \ref{app:p2o_prompt}.

\subsection{Prompt-to-Protocol (P2P)}
\label{sec:P2P (single loop)}

Unlike P2O (two loops), P2P is a single-loop scheme: the LLM directly proposes an executable charging protocol with an (arbitrary) explicit function, including its parameter values, based on a prompt containing various relevant pieces of information about desirable behaviour. 

\begin{algorithm}[ht]
\caption{P2P (single-loop) Direct Algorithm Generation}
\label{alg:P2P (single loop)}
\begin{algorithmic}[1]
\Require Initial pool $\mathcal{P}_0$ (set of $(s,L)$ pairs with pre-evaluated losses); 
         number of generations $T$; 
         children per generation $C$; 
         pool size $N$
\Ensure Best algorithm/protocol $s^\ast$
\State $\mathcal{P} \gets \mathcal{P}_0$
\For{$t = 1$ \textbf{to} $T$}
    \State $\mathcal{C} \gets \emptyset$ \Comment{initialize child set}
    \For{$c = 1$ \textbf{to} $C$}
        \State Select parents $p_1, p_2$ from $\mathcal{P}$ \Comment{e.g.\ tournament selection}
        \State $s_c \gets \textsc{LLMGenerate}(p_1, p_2)$ \Comment{LLM proposes a new protocol}
        \State $L_c \gets \textsc{Evaluate}(s_c)$ \Comment{simulation/experiment returns loss}
        \State $\mathcal{C} \gets \mathcal{C} \cup \{(s_c, L_c)\}$
    \EndFor
    \State $\mathcal{P} \gets \textsc{SelectTopN}(\mathcal{P} \cup \mathcal{C}, N)$ 
           \Comment{retain $N$ best candidates by loss}
\EndFor
\State \Return $s^\ast$ 
       \Comment{$(s^\ast, L^\ast) = \arg\min_{(s,L)\in \mathcal{P}} L$}
\end{algorithmic}
\end{algorithm}

Initially, a pool of candidate LLM-generated protocols is created and their loss is evaluated via experiment. At each subsequent generation, a subset of parent algorithms is selected from the current pool using tournament selection. The LLM is then prompted to generate new child candidates based on these selected parents. Each child algorithm is subsequently evaluated to obtain a performance metric. If a child outperforms any existing member of the pool, it replaces the worst-performing one, ensuring that the overall quality of the pool improves over time. This iterative cycle of selection, generation, evaluation, and replacement continues until the trial budget is exhausted or the population converges. The final pool thus constitutes a diverse yet high-performing set of candidate algorithms, from which the best protocol is selected. The complete procedure is summarized in Algorithm \ref{alg:P2P (single loop)}.

\subsection{Non-LLM benchmarks}
\label{sec:Non-LLM}
In the first case study, the performance of the P2O approach is compared against three more conventional gradient-free optimization strategies for the specification of the neural network architectures: Genetic Algorithms, Bayesian Optimization, and Random. In each case the solution space must be predefined by the user (e.g. variables to control the number of NN layers, variables to select activation functions from a list, variables to choose which normalizations to apply). This can be an arduous task and possibilities will often be missed. 

\subsection{Battery modeling}
\label{sec:pybamm}

We model battery behavior using a physics based Doyle–Fuller–Newman (DFN) electrochemical model augmented with degradation and thermal dynamics, implemented through PyBaMM \citep{sulzer2021python}. The DFN model resolves lithium diffusion in solid particles, electrolyte transport, charge transfer kinetics, and voltage response under arbitrary current profiles, allowing protocol dependent aging phenomena such as SEI growth, loss of active material, and particle cracking to be captured. To reflect realistic cycling conditions, we couple the electrochemical model to a lumped thermal model that accounts for ohmic, kinetic, and entropic heat generation as well as heat dissipation to the environment. Degradation is simulated using reaction limited SEI growth with porosity change, stress driven electrode damage, and cracking submodels. Because full lifetime simulation is computationally expensive, we accelerate aging by increasing selected degradation rate parameters while verifying that protocol rankings remain consistent under non accelerated conditions.

\section{Results}
We conducted three case studies with progressively increasing complexity, allowing us to rigorously validate each component of the proposed framework before demonstrating its practical utility. Case 1, the predefined-protocol search, serves as a controlled and simplified setting to test the effectiveness of the outer-loop optimization in P2O. Case 2, the constant-heating protocol search, introduces a non-differentiable setting and focuses on validating the performance of the inner-loop SAASBO optimizer, while also providing a direct comparison with P2P. Building on these foundations, Case 3 presents a more realistic and application-oriented scenario: discovering an adaptive fast-charging protocol. All experiments are conducted using GPT-5 as the LLM. In the case studies, we used PyBaMM\citep{sulzer2021python} to simulate charging protocols in order to rapidly validate our method with a well-established battery model. In real applications, the same optimization framework can be applied by simply replacing the PyBaMM simulation with charge–discharge experiments on real batteries, although this would of course incur greater expense.

\subsection{Case 1: Predefined heat search}

We begin with a deliberately simplified setting to isolate and evaluate the effectiveness of the outer-loop neural architecture optimization in the P2O (two-loop) framework. The predefined targets in this case are not meant to represent realistic charging protocols; rather, they create a controlled environment where the influence of the outer loop can be examined without confounding factors. To eliminate interference from the inner loop, we adopt a stable and fully differentiable setup in which the inner-loop neural network parameters are optimized using standard gradient descent.

\begin{figure*}[h]
\centering
    \includegraphics[width=1\linewidth]{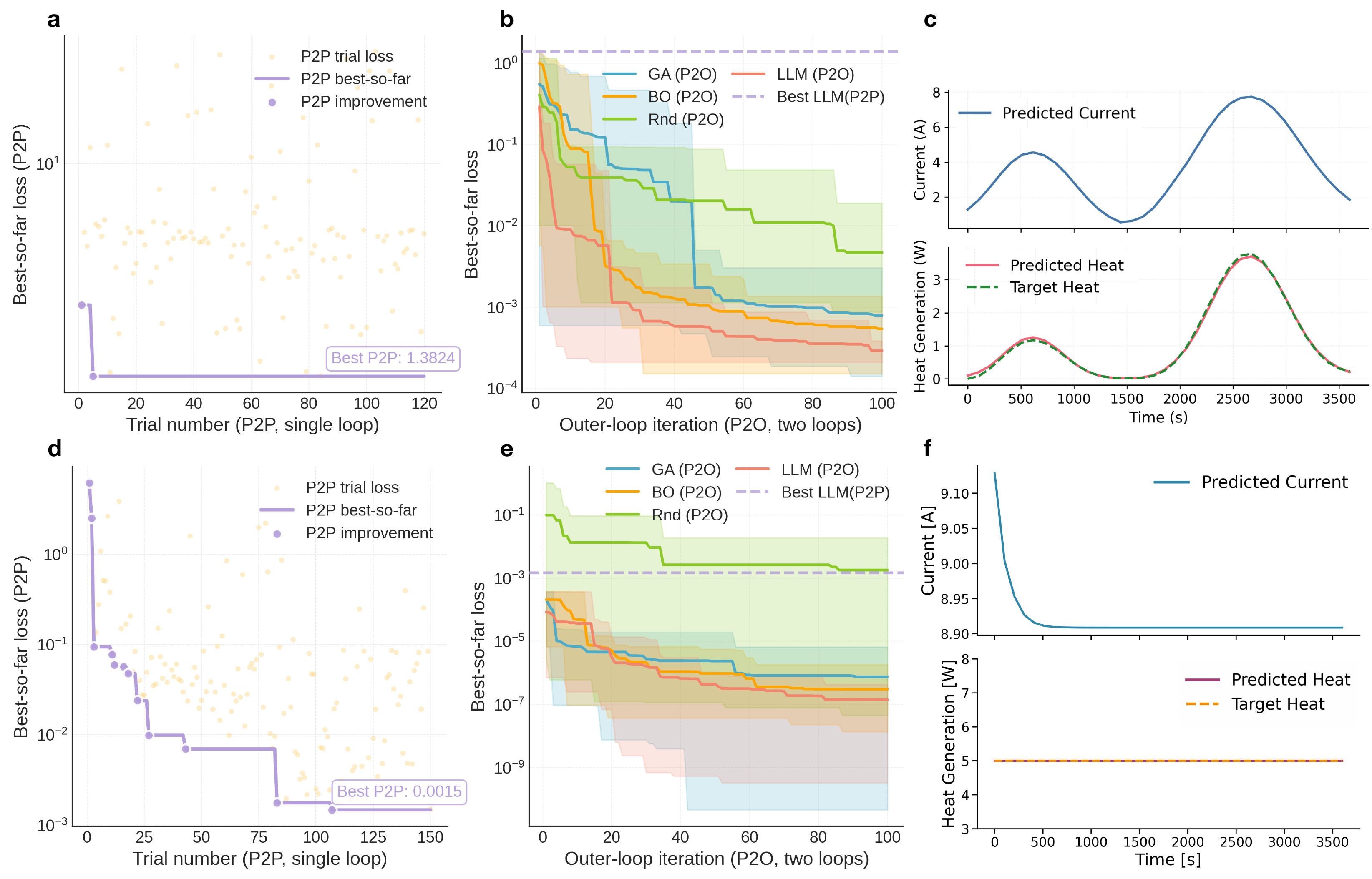}
    \caption{Optimization results for P2O (two loops) and P2P (single loop) across two tasks. (a–c) Complex predefined protocol; (d–f) constant-heating protocol. (a, d) show P2P performance, where LLM directly generates explicit charging functions. (b, e) compare P2O variants using different outer-loop mechanisms—LLM-guided, BO, GA, and Random Sampling; each curve shows the cumulative best loss over iterations, where the solid line denotes the mean across 10 runs and the shaded area marks the range between the minimum and maximum best losses; the LLM-guided approach converges fastest and achieves the lowest loss (dashed lines mark best P2P results). (c, f) display predicted current and heat generation profiles of the best-performing protocols.
    }
    \label{fig:eval_outer}
\end{figure*}

Under this simplified setting, the task is to recover two predefined profiles - a complex explicit function and a simple constant-heating pattern. This allows us to assess how effectively the outer loop can guide architecture evolution. To contextualize the performance of P2O, we compare it against common baselines such as Bayesian Optimization (BO), Genetic Algorithm (GA), and Random Search. We also include a preliminary comparison with P2P (single loop), which, unlike P2O, is entirely gradient-free since it directly generates explicit protocols without an inner-loop optimizer.

Since Case 1 aims only to examine outer-loop optimization rather than full physical fidelity, we use a simple Equivalent Circuit Model (ECM) combined with a lumped thermal model (see Appendix~\ref{appendix:ecm}) to evaluate protocol performance and enable gradient-based training of the neural network applied to a non-linear system.  Model performance is measured using the mean squared error (MSE) between the predicted and target heat profiles:

\begin{equation}
L = \frac{1}{N}\sum_{i=1}^{N}(Q_{\text{pred},i} - Q_{\text{target},i})^2
\end{equation}

In the LLM-guided variant, LLM initially generates ten diverse seed architectures and then three offspring architectures per seed at each iteration to promote exploration (prompt templates in Appendix~\ref{app:p2o_prompt}). The LLM is encouraged to explore broadly rather than follow fixed layer-width patterns. To maintain compactness, it is explicitly instructed to propose lightweight models, and any design exceeding 35 trainable parameters is discarded before evaluation. The BO, GA, and Random Sampling variants use Optuna-guided sampling \citep{akiba2019optuna}, a standard genetic algorithm, and independent random proposals, respectively. Their search space is restricted to compact multilayer perceptrons (MLPs) with at most 3–4 hidden layers, 1–5 neurons per layer, varied activation functions, and optional normalization layers. 

For all four variants, each candidate network maps 1000 uniformly sampled time steps (1–3600 s) to a 1000-step current profile, which is then evaluated using a battery simulation model to compute heat generation. The 35-parameter upper bound is applied uniformly across all variants (see Appendix~\ref{app:setup}). 

P2P (single-loop) operates without an inner optimization loop. Instead of training, it directly prompts the LLM to generate an explicit time-dependent function \( I(t) \) that serves as the charging protocol. The prompt (see Appendix~\ref{app:p2p_prompt}) specifies only the desired protocol behavior and the constant-heat requirement, without exposing any battery-model details. The LLM produces ten diverse seed functions, along with three offspring variants in each iteration. All generated functions are evaluated directly, with no further optimization.

In both predefined explicit function and constant heating rate experiments, the LLM-guided P2O (two-loop) framework performs strongly in the outer loop as shown in Figure~\ref{fig:eval_outer}. For the complex predefined function task, the LLM-guided search reduces loss by nearly two orders of magnitude within 20 iterations, outperforming baselines after five evaluations. This early advantage likely reflects its ability to propose architectures informed by prior knowledge. BO shows comparable early progress but higher variance, EA improves slowly with a single large gain near iteration 45, and Random search remains one order of magnitude worse. In contrast, performance differences are less pronounced in the simpler constant-heating case. Because the target profile is structurally simple and imposes minimal demands on network architecture, EA, BO, and the LLM-guided approach all perform competitively. BO shows stable convergence but can become trapped in local minima, whereas both EA and the LLM reach similarly low loss values.

In complex predefined function tasks, P2P (single-loop) performs notably worse than P2O (two-loop). However, under simple constant-current protocols, P2P achieves a best loss of about \(1.5\times10^{-3}\), comparable to a gradient-trained neural network—remarkable for a gradient-free method. Although P2O variants reach lower losses, they operate in a different regime, using gradient-based inner-loop optimization and requiring far fewer simulation runs. These results suggest that P2P performs well on simple, regular protocols but struggles with irregular ones when the prompt does not provide clear or directly useful structural guidance. Detailed comparisons are presented in the next section.

Notably, the design space explored by the LLM-guided variant is fundamentally inaccessible to traditional algorithms such as genetic algorithms or random sampling. These methods require a predefined, human-specified search space, typically a limited set of layer counts, widths, or activation patterns. Exhaustively expanding this space to match the flexibility of LLM-generated architectures would dramatically increase dimensionality and make optimization substantially more difficult. More philosophically, any search space defined by humans inevitably reflects our cognitive biases and modeling limitations. Once we commit to a particular parametrization, we compress the problem into a form we are capable of specifying, often simplifying or distorting its underlying complexity. In contrast, LLMs are not constrained by such manually imposed structures: by generating architectures directly through language, they can explore broad, expressive, and loosely constrained design spaces that extend beyond explicit parameterization. This ability to operate outside human-designed search boundaries represents a key advantage of the LLM-based approach.

The optimal neural architectures obtained in the constant-heating setting differ in structure, ranging from relatively complex multi-branch networks to simpler feed-forward layers and compact residual units with smooth nonlinearities. The first two architectures correspond to larger search spaces. In contrast, the last one induces a much smaller family of protocol shapes; however, its typical forms closely match the optimal protocol’s decreasing–then–flattening pattern (Fig.~\ref{fig:eval_outer}f), which in turn enables strong optimization performance. Detailed visualizations of the associated search spaces are provided in Appendix~\ref{app:Case1_P2O_nn} (Fig.~\ref{fig:search_space}).


\subsection{Case 2: Constant heat protocol search}
In practice, battery research relies on experimental validation, which is inherently non-differentiable. This case study therefore considers a non-differentiable setting to evaluate the inner-loop optimization of P2O (two loops) and compare it with P2P (single loop). The study aims to identify a charging protocol that maintains a constant heat generation rate of 0.4 W over one-hour charge, employing a Doyle–Fuller–Newman (DFN) model coupled with a lumped thermal model. The mean squared error (MSE) between the predicted and target heat serves as the evaluation metric.


A compact neural network identified from Case Study 1 is used for testing the inner loop. The neural network's parameters are optimized using SAASBO and Random search, each repeated ten times. The P2P (single loop) method is also tested, where the LLM directly evolves explicit charging protocols. All approaches are compared under equivalent battery simulation costs or trial counts, which dominate computational expense in practical scenarios.

\begin{figure*}[h]
\centering
    \includegraphics[width=1\linewidth]{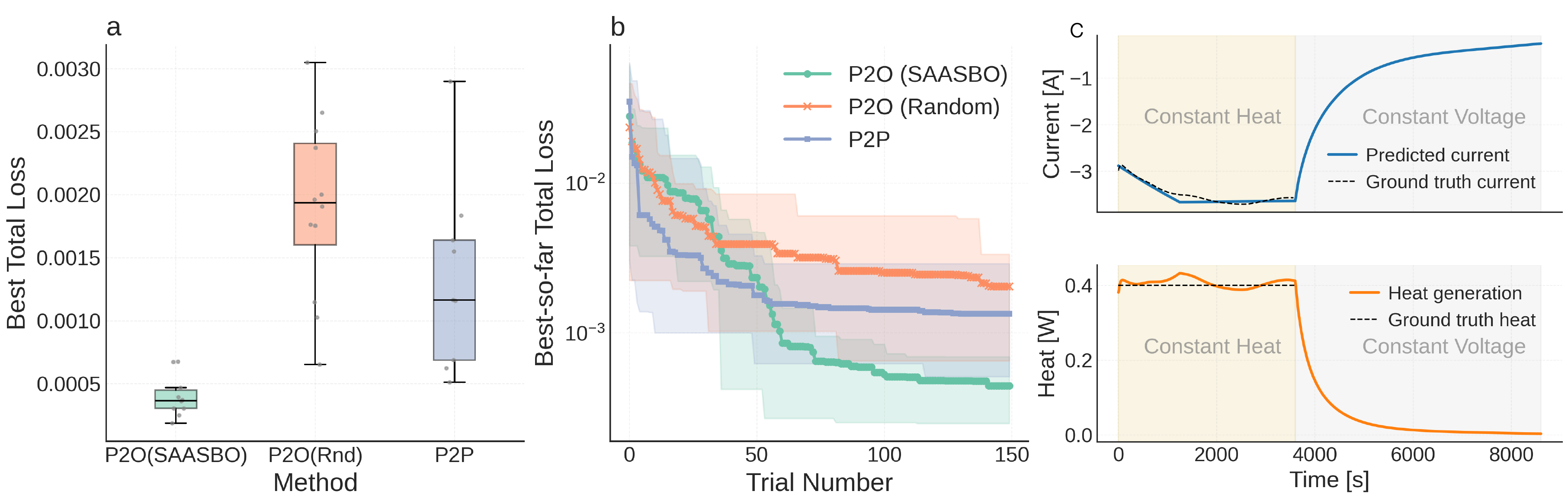}
    \caption{Optimization under a gradient-free setting.  
(a) Best final loss across ten runs of SAASBO and Random Search (inner-loop optimizers of P2O) and the LLM-based P2P method (single-loop).  
(b) Convergence of the best-so-far loss over battery-simulation trials for all three approaches.  
(c) Verification of the optimized charging protocol obtained from SAASBO.
    }
    \label{fig:eval_inner}
\end{figure*}

As shown in Figure~\ref{fig:eval_inner}, P2O using SAASBO consistently outperforms random search, demonstrating superior convergence stability. The P2P results fall between those of SAASBO and random search, which is expected because SAASBO is a dedicated parameter optimization algorithm, whereas the LLM in P2P relies on heuristic reasoning based on linguistic cues. Although the prompt did not include explicit physical guidance on how to maintain a heat generation rate of 0.4 W, P2P still achieved effective optimization within 150 trials, with particularly rapid improvement during the first 50 iterations.

 The best-performing protocol generated by the LLM in the P2P approach is shown in Appendix~\ref{app:case2_p2o_alg}. When asked to explain its proposed method, the LLM indicated that the approach was not derived from existing literature or any clear physical principle. Upon closer inspection, this protocol does not exhibit an obvious physical rationale but instead appears to be a problem-specific heuristic, refined through iterative adjustments to both parameters and structural components during the optimization process. Overall, these results suggest that when the prompt lacks information that can be directly utilized for algorithm design, P2O (two loops) provides more reliable optimization through dedicated parameter tuning, albeit at the cost of additional evaluations. However, it should be noted that P2P does initially perform well.

To validate whether our optimization algorithm has learned a near-optimal charging protocol, we compute the reference current (based on PyBaMM's differential stepping) that exactly maintains 0.4 W of heat generation and compare it with the optimized result (N.B. this reference protocol is not used during optimization). As shown in Fig.~\ref{fig:eval_inner}c, the optimized current closely matches the constant-heat reference, demonstrating that the learned protocol captures the desired thermal behavior. Naturally, this ground truth analysis cannot be performed on real experiments or under more complex simulation conditions.

\subsection{Case 3: Adaptive fast charging protocol search}
\begin{figure*}
\centering
    \includegraphics[width=1\linewidth]{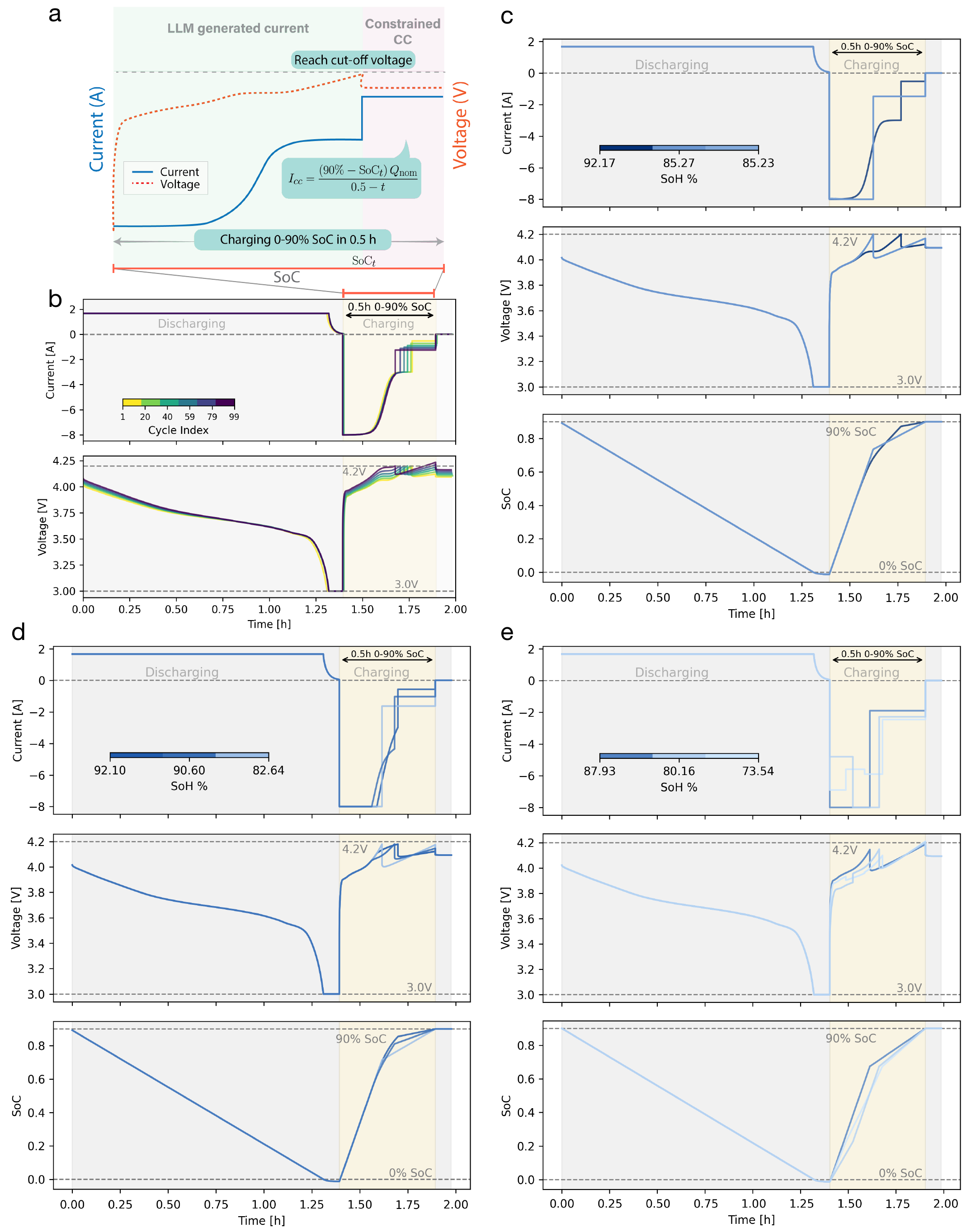}
    \caption{Comparison of LLM-generated charging protocols with conventional baselines. (a) Charging-phase schematic: the LLM generates a variable-current profile until the cut-off voltage, followed (if needed) by a constrained CC phase to reach the target SOC within the time limit. (b--e) Full charge--discharge trajectories; the yellow region marks the charging phase (LLM segment + constrained CC). (b) Best P2O protocol across cycles, showing earlier voltage-limit entry with degradation. (c) P2O, (d) P2P, and (e) CCCV: best result with representative best, worst, and intermediate trajectories.}
    \label{fig:nn_cccv}
\end{figure*}

This case study provides a more realistic simulation of practical charging scenarios. The objective is to identify a fast-charging strategy that minimizes battery degradation over the same number of charge cycles. To ensure a fair comparison among different charging protocols, we follow the principle used in the state-of-the-art work by Attia \textit{et al.} \citep{attia2020closed}, by fixing the total charging time required to reach a target state of charge (SOC). In their work, this was achieved using a protocol composed of three constant-current (CC) segments followed by a constrained CC phase. In contrast, our method uses either an LLM-generated neural network to produce a flexible current profile or an LLM-generated direct protocol that enables a broader exploration of the search space. Once the cell voltage approaches the cut-off threshold, a constrained CC phase is applied to ensure that the target SOC is achieved within the allotted charging time (see Fig.~\ref{fig:nn_cccv}a).

In real charging scenarios, a battery may begin at different voltages, temperatures, and states of charge, rather than from a fixed initial condition. Therefore, we adopt an adaptive charging scheme in which the neural network or explicit function maps the instantaneous voltage, temperature, and SOC to the output current. This setup differs from Case Study 1, where the charging protocol is a direct time-to-current mapping. At each time step, the network or algorithm receives the current battery state and outputs the charging current, which is then applied to the battery model; the updated state at the next step is used as input for subsequent predictions. The neural network architecture is flexible and may take various forms, including multilayer perceptrons (MLPs), recurrent neural networks (RNNs), or other lightweight structures generated by the LLM.


In this case study, each simulation cycle consists of five stages:  
(i) discharge at a constant current of $C/3$ until the cell voltage reaches $2.5\,\mathrm{V}$;  
(ii) voltage hold at $2.5\,\mathrm{V}$ until the current decreases below $50\,\mathrm{mA}$;  
(iii) neural-network-controlled charging, where the current is given directly by the neural network as a function of the instantaneous voltage, temperature, and SOC, and applied until one of the following conditions is met: $V = 4.20\,\mathrm{V}$, $\mathrm{SOC}=90\%$, or $t = t_{\max} = 0.5\,\mathrm{h}$;  
(iv) constrained CC charging (if required), where, if $\mathrm{SOC}<90\%$ and time remains, the constant current is set to  
\begin{equation}
I_{\mathrm{cc}} = -\frac{(\mathrm{SOC}_{\mathrm{tgt}} - \mathrm{SOC})\,Q_{\mathrm{nom}}}{t_{\mathrm{rem}}}
\end{equation}
and applied until either the target SOC is reached; otherwise, if the time limit is reached first, the protocol is discarded. Here $I_{\mathrm{cc}}$ is the imposed constant charging current, $\mathrm{SOC}_{\mathrm{tgt}} = 90\%$ is the target state of charge, $\mathrm{SOC}$ is the instantaneous state of charge at the start of this stage, $Q_{\mathrm{nom}}$ is the nominal cell capacity, and $t_{\mathrm{rem}}^{h}$ is the remaining charging time;  
(v) finally, a rest period of $300\,\mathrm{s}$ at zero current is imposed. The procedure is detailed in Algorithm~\ref{alg:battery-simulation}.

\begin{algorithm}[ht]
\caption{Battery Cycling with LLM Derived Charging Policy}
\label{alg:battery-simulation}
\begin{algorithmic}[1]
\Require nominal capacity $Q_{\text{nom}}=2.5\,\mathrm{Ah}$;
target $\mathrm{SOC}_{\text{tgt}} = 90\%$;
maximum charging time $t_{\max} = 0.5\,\mathrm{h}$;
voltage threshold $V_{\text{thr}} = 4.20\,\mathrm{V}$;
rest time $300\,\mathrm{s}$;
LLM derived charging policy $\pi$ that maps $(V, T_{\text{cell}}, \mathrm{SOC})$ to current $I$
\For{$c=1$ to $N$}

  \State \textbf{(A) Discharge:} $I \gets Q_{\text{nom}}/3$ until $V \le 2.5\,\mathrm{V}$

  \State \textbf{(B) CV tail:} hold $V \gets 2.5\,\mathrm{V}$ until $|I| \le 0.05\,\mathrm{A}$

  \State \textbf{(C) Policy controlled charge:} $t \gets 0$
  \While{$t < t_{\max}$ \textbf{and} $V < V_{\text{thr}}$ \textbf{and} $\mathrm{SOC} < \mathrm{SOC}_{\text{tgt}}$}
    \State $I \gets \pi(V, T_{\text{cell}}, \mathrm{SOC})$
    \State Apply $I$ for $\Delta t$ to update $V$, $T_{\text{cell}}$, and $\mathrm{SOC}$
    \State $t \gets t + \Delta t$
  \EndWhile

  \State \textbf{(D) CC constrained stage (if needed):}
  \If{$\mathrm{SOC} < \mathrm{SOC}_{\text{tgt}}$ \textbf{and} $t < t_{\max}$}
    \State $t_{\text{rem}}^{h} \gets (t_{\max} - t)/3600$
    \State $I_{\text{cc}} \gets -\dfrac{(\mathrm{SOC}_{\text{tgt}} - \mathrm{SOC}) Q_{\text{nom}}}{t_{\text{rem}}^{h}}$
    \While{$t < t_{\max}$ \textbf{and} $\mathrm{SOC} < \mathrm{SOC}_{\text{tgt}}$}
      \State Apply $I_{\text{cc}}$ for $\Delta t$ to update $V$, $T_{\text{cell}}, \mathrm{SOC}$
      \State $t \gets t + \Delta t$
    \EndWhile
  \ElsIf{$\mathrm{SOC} = \mathrm{SOC}_{\text{tgt}}$ \textbf{and} $t \le t_{\max}$}
    \State skip CC stage
  \Else
    \State discard cycle
  \EndIf

  \State \textbf{(E) Rest:} $I \gets 0$ for $300\,\mathrm{s}$

\EndFor
\State record SOH

\end{algorithmic}
\end{algorithm}

In this case study, we used PyBaMM to simulate the effect of the charging protocols, based on the DFN model plus degradation models. The cell was parameterized using the \texttt{Ai2020} dataset \citep{ai2020electrochemical} and equipped with the following degradation submodels: swelling-induced cracking \citep{deshpande2012battery}, stress-driven loss of active material \citep{ reniers2019review}, reaction-limited SEI growth with porosity change \citep{yang2017modeling}, and SEI formation on cracks \citep{deshpande2012battery}. A lumped thermal model was also included, and all degradation and thermal processes were integrated within the DFN framework as detailed in Appendix~\ref{app:case3_model}.

\begin{figure*}[h]
\centering
    \includegraphics[width=0.8\linewidth]{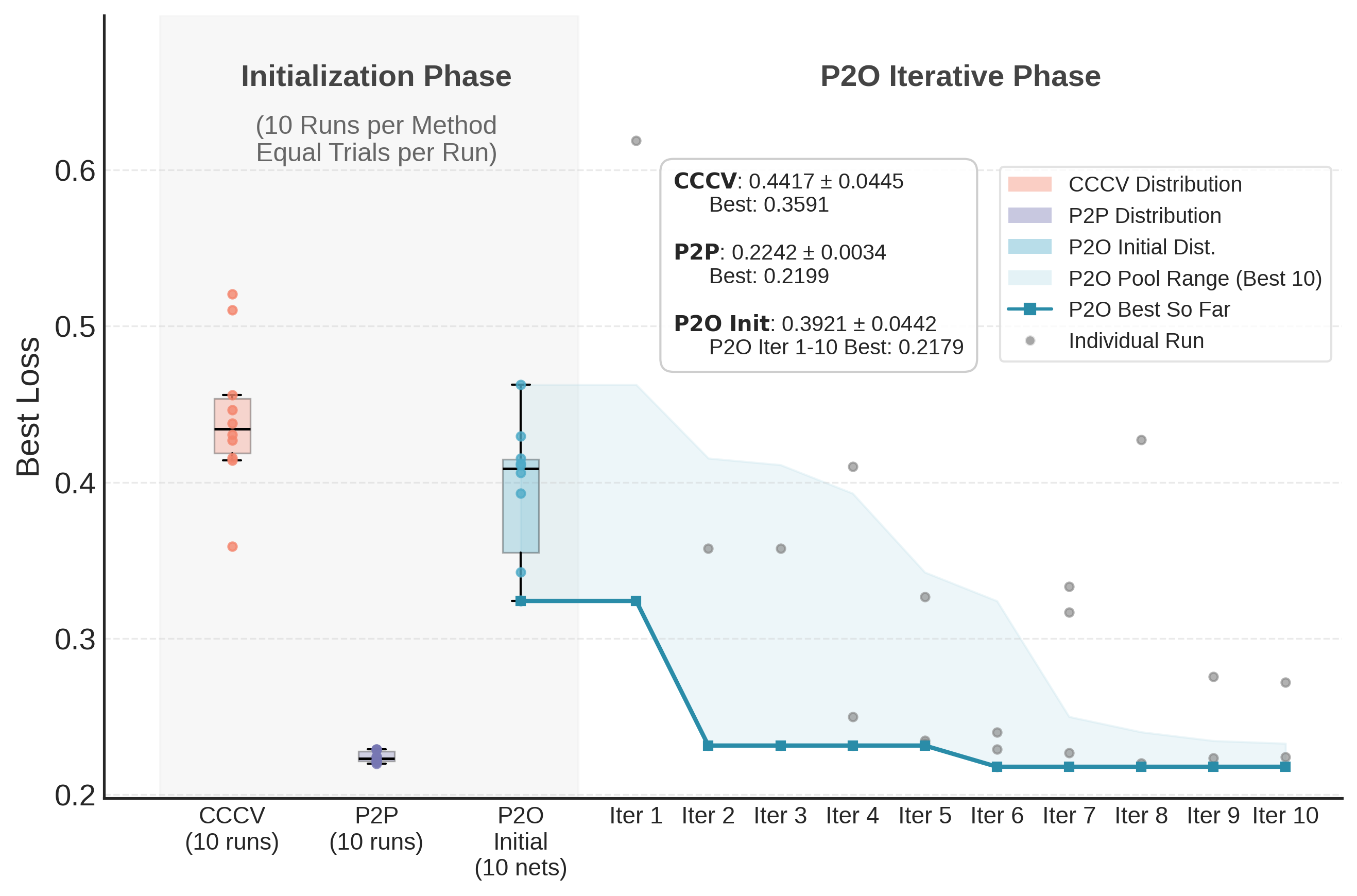}
    \caption{Best loss comparison under controlled computational budgets. The multi-step CC baseline (CCCV) and P2P method were evaluated over 10 independent runs, each limited to 210 optimization attempts to match the inner-loop budget of the P2O method. The left panel shows the variance in best loss for 10 CCCV runs, 10 P2P runs, and P2O initialization (10 nn structures). The right panel demonstrates the iterative improvement of P2O over 10 outer-loop iterations (with 210 inner SAASBO evaluations per iteration). Note that P2P performs only algorithmic iterations without an inner-loop parameter optimization, whereas CCCV involves only parameter optimization without an outer-loop iterative process; therefore, only P2O exhibits a full iterative trajectory in the right panel.
    }
    \label{fig:case3_com}
\end{figure*}
                
To accelerate aging and obtain results within a feasible runtime, we increased the degradation rate by modifying several model parameters, most notably by scaling the SEI reaction exchange current density by a factor of five. Additional adjustments were made to electrode loss terms, particle cracking rates, the heat transfer coefficient, and the ambient temperature (see Appendix~\ref{app:case3_acc}). In the simulation, we ran 100 charge–discharge cycles under this accelerated aging setup. To verify that this accelerated aging scheme preserves the relative ranking of charging protocols, we simulated three representative protocols for 1000 cycles without acceleration. The results showed that the relative performance ordering remained consistent across both settings (Appendix~\ref{app:no_acc}). Because all charging protocols must reach the target SOC within a fixed 0.5-hour window, occasional voltage overshoot is unavoidable. Although the degradation model captures some voltage-driven processes, it does not include all harmful side reactions. We therefore introduce an additional voltage penalty term to account for these unmodeled effects and to promote voltage-safe charging strategies (Appendix~\ref{app:voltage_penalty}). At the end of each cycle, the state-of-health (SOH) is computed as a composite measure combining the dischargeable capacity at the end of cycle $i$, denoted $Q_{\mathrm{cap}}^{(i)}$, with the capacity-equivalent voltage penalty denoted $\mathcal{P}^{(i)}$. The SOH is defined as
\begin{equation}
\mathrm{SOH}^{(i)}
= \frac{Q_{\mathrm{cap}}^{(i)}}{Q_{\mathrm{nom}}}
-
\frac{\mathcal{P}^{(i)}}{Q_{\mathrm{nom}}}
\end{equation}
with $Q_{\mathrm{nom}}$ representing the nominal cell capacity.

We define the loss function based solely on the SOH after the final cycle. If the final SOH drops below a threshold, a large constant penalty $C_{\mathrm{penalty}}$ is assigned to reject such solutions. Otherwise, the loss decreases smoothly with increasing SOH. Formally, the loss function is defined as:

\begin{equation}
L = 
\begin{cases}
C_{\mathrm{penalty}}, & \text{if } \mathrm{SOH}^{(N)} < 0.6 \\
-\ln\left( \dfrac{\mathrm{SOH}^{(N)} - 0.6}{1 - 0.6} \right), & \text{if } \mathrm{SOH}^{(N)} \geq 0.6
\end{cases}
\end{equation}

In our work, SOC is defined by Coulomb counting, i.e., by integrating the charging current and normalizing by the nominal capacity $Q_{\text{nom}}$:  
\begin{equation}
\mathrm{SOC}(t) = \mathrm{SOC}(0) - \frac{1}{Q_{\text{nom}}} \int_{0}^{t} I(\tau)\, d\tau 
\end{equation}
where $I(\tau)$ is the applied current at time $\tau$. This definition is consistent with experimental practice, where SOC cannot be measured directly and is always inferred from the accumulated charge. In each cycle, the cell is first fully discharged (thereby resetting the SOC to zero) and then charged to a fixed target SOC. We set this target to 90\% rather than 100\% for two considerations. First, as the cell ages and its usable capacity declines, a nominal 100\% SOC (defined with respect to the initial rated capacity) no longer corresponds to a true full charge. Therefore, using a slightly reduced target avoids artificially pushing the cell into an overcharged state as capacity fades. Second, terminating the charge below full capacity is a well-established practice in fast-charging protocols, as it limits exposure to the high-voltage regime and thus alleviates voltage-induced degradation.


To provide a comparative baseline, we implemented the state-of-art method proposed by \cite{attia2020closed}. Specifically, we replaced our LLM derived current profile with a multi-step constant current (CC) charging approach, using three consecutive CC segments terminating at 20\%, 40\%, and 60\% SOC, consistent with the procedure described in their paper. This approach maintains the same 0.5-hour duration to reach an 90\% SOC as in our method.

The multi-step CC baseline contains only fixed-current segments and therefore requires optimization solely in the inner loop. To ensure a fair comparison, both P2P (which has only an outer loop) and the multi-step CC baseline were allocated the same number of simulation evaluations: 210 optimization attempts per run, repeated for 10 independent runs. P2O includes both an outer loop and an inner-loop Bayesian optimization. During the initialization phase, we sampled 10 random neural networks, matching the initial evaluation count of the other methods. In the evolutionary stage, we ran 10 outer-loop iterations, and in each iteration the inner loop conducted 210 SAASBO evaluations. Figure \ref{fig:case3_com} presents the resulting performance comparison. Fig.~\ref{fig:nn_cccv} summarizes the results. Panels (c-d) show the best protocol from each method, selected from all runs, together with several representative examples. Panel (b) depicts the full 100-cycle trajectory of the top-performing LLM-generated protocol discovered under the P2P configuration.

As shown in Fig.~\ref{fig:nn_cccv}, both P2P and P2O substantially outperform the multi-stage CCCV baseline, improving final SOH by 4.24\% and 4.17\%, respectively. In the Initialization Phase of Fig.~\ref{fig:case3_com}, all methods were given the same number of experimental evaluations, ensuring a fair comparison. Under this equal-trial setting, P2P achieves the best and most stable performance, while P2O already exceeds the CCCV baseline even before any outer-loop evolution. P2P directly reaches its final performance within this initial trial budget, whereas P2O attains its best result only after additional iterations. The strong performance of P2P can be attributed to the structure of its prompt: it explicitly states that exceeding 4.2 V incurs a penalty. This allows the LLM to internalize voltage-safety requirements and naturally avoid unsafe regions, consistently generating voltage-compliant protocols. Such constraint encoding through language is a unique advantage of LLM-based optimization. In contrast, the neural-network parameterization used in P2O cannot express constraints linguistically and must learn these behaviors indirectly through repeated trial-and-error in the inner loop. This difference explains why P2P matches P2O’s final performance without requiring an inner loop and with far fewer experimental trials.

The original multi-step CC baseline includes only three parameters, which makes it considerably less expressive than our method. To narrow this gap, we increased the number of CC segments to 35, thereby providing the baseline method with a comparable level of flexibility. However, this high-dimensional parameterization of a piecewise function proved extremely difficult to optimize; under the same number of optimization attempts with SAASBO, the method was unable to produce competitive results. The resulting SOH trajectories exhibited pronounced sawtooth patterns, as shown in Appendix Fig.~\ref{app:35cc}. This behavior highlights an inherent trade-off in multi-step CC parameterizations: while too few segments overly restrict the solution space, too many segments becomes impossible to optimize, yielding jagged control signals.

\section{Discussion and Conclusion}
Optimizing the shape of battery charging protocols is inherently challenging due to the high cost of simulation-based evaluation. As a result, many existing methods restrict the search space, which limits both flexibility and performance. Our study demonstrates that LLM-driven closed-loop optimization offers a promising alternative. The two proposed methods, P2O and P2P, each improve the final SOH by approximately 4.2\% over state-of-the-art protocol optimization techniques, indicating that integrating language models into the optimization pipeline can meaningfully extend the reachable design space.

The two approaches, however, differ in their strengths. P2O (Prompt-to-Optimization; two loops) is well suited for complex or high-dimensional search spaces: the LLM proposes protocol structures, and an inner optimizer tunes their parameters. This broad exploration capability comes at higher computational cost. P2P (Prompt-to-Protocol; single loop) directly generates full protocols from language prompts, making it far more efficient. When the prompt provides clear constraints such as limits on voltage, P2P can exploit the LLM’s semantic understanding to produce high-quality protocols with far fewer trials. When the prompt lacks explicit structural guidance, P2O offers a more reliable alternative that delivers better results.

Although our case studies rely on simulation for rapid validation, both methods are gradient-free and therefore directly applicable to real experiments. When simulations are reasonably accurate, optimization can first be performed \textit{in silico} and then fine-tuned on physical batteries. For P2O, simulation helps identify promising neural-network architectures for the outer loop, after which only the inner-loop optimization needs to be run experimentally, potentially augmented with multi-fidelity Bayesian optimization to combine simulation and real data efficiently. For P2P, simulation can be used to update the protocol pool, enabling real experiments to continue improving protocols from a simulation-informed starting point. These pathways show that LLM-driven optimization is not only effective in simulation but also naturally suited for efficient deployment in real-world battery-protocol design.

This work highlights the potential of large language models as powerful tools for optimization. LLM-based optimization can remain effective even when each evaluation is expensive, without requiring thousands of trials. LLMs allow constraints and design preferences to be expressed directly in natural language - often far more easily than through mathematical specification, and in some cases in ways that would be nearly impossible to encode otherwise. Their generative capacity enables exploration beyond predefined search spaces, and their broad prior knowledge supports more informed and efficient initialization. It should be noted that we had to re-run case study 1 several times as the LLM specific network architecture features that we had not previously through to include in the search space for the evolutionary algorithm. 

In future work, we plan to incorporate rich, language-based feedback rather than relying solely on scalar rewards, for example, description of the issues observed in the charging protocol (e.g. the location of unwanted voltage spikes). Future extensions could also include voltage-control concepts, such as dynamically targeting internal electrochemical or lithium-plating overpotentials rather than terminal voltage alone, to further accelerate charging while maintaining safety margins. We hope that the framework presented here offers a promising new direction not only for battery-protocol optimization, but also provides insights for other domains where experiments are slow, expensive, and consequently challenging to optimize.

\section*{Code Availability}
The code needed to reproduce the results of the paper/use the tool is a available at     \url{https://github.com/tldr-group/prompt-to-protocol}. 

\section*{Acknowledgements}
This work was supported by the Imperial Lee Family Scholarship awarded to G.L. We would like to thank the members of the TLDR group for their valuable comments and insightful discussions.

\bibliographystyle{plainnat}  
\bibliography{references}


\newpage

\appendix
\renewcommand{\thefigure}{\thesection.\arabic{figure}}
\renewcommand{\thetable}{\thesection.\arabic{table}}

\makeatletter
\@addtoreset{figure}{section}
\@addtoreset{table}{section}
\makeatother

\section{SAASBO}

\begin{algorithm}[ht]
\caption{\textsc{SaasboOptimize}$(s,\,B)$}
\label{alg:saasbo_opt}
\begin{algorithmic}[1]
\Require Network structure $s$ ; BO budget $B$
\Ensure  Best params $\theta^\ast$ and loss $L^\ast$
\State Initialise SAAS-GP surrogate with sparse axis-aligned prior
\State Draw an initial Sobol/LHS batch, evaluate $\{(\theta_i,L_i)\}_{i=1}^{n_0}$
\For{$k = 1$ to $B$}
    \State Sample GP hyper-parameters via MCMC  \Comment{fits sparse subspace}
    \State $\theta_k \gets \arg\max_{\theta}$ Acquisition$(\theta\mid\text{SAAS-GP})$
    \State $L_k \gets \textsc{TrainAndEval}(s,\theta_k)$
    \State Update SAAS-GP with $(\theta_k,L_k)$
\EndFor
\State $(\theta^\ast,L^\ast)\gets\min_{\;(\theta,L)} L$
\Return $\theta^\ast,\ L^\ast$
\end{algorithmic}
\end{algorithm}

\section{Prompts}

\subsection{P2P Prompt}
\label{app:p2p_prompt}
\noindent
Note: Each case below introduces slight variations in the prompt configuration to test the model’s ability to adapt the algorithm generation task under different physical and operational conditions.

\subsubsection*{P2P — Case 1}
Add to the first line in the constant heating experiment:
\begin{tcolorbox}[promptbox]
to maintain constant heat generation rate.
Battery current range: 0–10 A.
\end{tcolorbox}

Add to the first line in the explicit function experiment:
\begin{tcolorbox}[promptbox]
to match the predefined protocol.
Battery current range: 0–10 A.
\end{tcolorbox}

\subsubsection*{P2P — Case 2}
Add to the first line:
\begin{tcolorbox}[promptbox]
to maintain constant heat generation rate as 0.4 W.
Battery current range: 0–5 A.
\end{tcolorbox}

\subsubsection*{P2P — Case 3}
Add the following Battery Specifications:
\begin{tcolorbox}[promptbox]
The voltage range should be between 3.0 V and 4.25 V, with an additional penalty applied for operation above 4.20 V.  
Temperature starts at 308.15 K.  
Maximum allowed charging current: 8 A.  
Minimum allowed charging current: 3 A.  
Charging time should be close to 0.5 hours.
\end{tcolorbox}

\subsubsection*{Base Prompt for P2P Initialization}
\begin{tcolorbox}[promptbox]
Generate 10 battery charging control algorithms using PyBaMM with the following constraints:

\begin{enumerate}[nosep]
  \item Battery specifications: maximum allowed charging current 10 A; minimum allowed charging current 0 A.
  \item The charging current must be negative (e.g., \texttt{-5} means charging at 5 A); in the simulation, negative current indicates charging.
  \item The absolute value of the charging current must not exceed 10 A.
  \item Do not use any Boolean operators or logic statements such as \texttt{if}, \texttt{and}, \texttt{or}, \texttt{not}, or similar.
  \item Use PyBaMM functions and variables only: \texttt{pybamm.t} (time variable), \texttt{pybamm.sin}, \texttt{pybamm.cos}, \texttt{pybamm.exp}, \texttt{pybamm.tanh}, \texttt{pybamm.sqrt}, etc.
  \item Do not use \texttt{pybamm.pi}; if you need pi, use \texttt{3.14} instead.
  \item Try different algorithms; do not use the same or highly similar algorithm multiple times.
  \item Each function must be named \texttt{current\_function} and must not take any parameters.
  \item Each function must return a PyBaMM expression that represents the charging current as a function of time.
  \item Make sure every Python code block is enclosed by matching triple backticks and starts with \texttt{python}.
\end{enumerate}

Below is the template for each charging function:

\begin{verbatim}
# Charging Function 1
```python
def current_function():
    t = pybamm.t
    # Your algorithm here: build a PyBaMM expression i(t)
    return i  # i is the charging current expression
```
\end{verbatim}

Repeat for Charging Function 2 through 10.
\end{tcolorbox}

\subsubsection*{Base Prompt for P2P Evolution Tasks}

\begin{tcolorbox}[promptbox]
Generate a battery charging control algorithm based on the two given algorithms to maintain a constant heat generation rate using PyBaMM with the following constraints:

\begin{enumerate}[nosep]
  \item Battery specifications: maximum allowed charging current 10 A; minimum allowed charging current 0 A.
  \item The charging current must be negative (e.g., \texttt{-5} means charging at 5 A); in the simulation, negative current indicates charging.
  \item The absolute value of the charging current must not exceed 10 A.
  \item Do not use any Boolean operators or logic statements such as \texttt{"if"}, \texttt{"and"}, \texttt{"or"}, \texttt{"not"}, or similar.
  \item Use PyBaMM functions and variables only: \texttt{pybamm.t} (time variable), \texttt{pybamm.sin}, \texttt{pybamm.cos}, \texttt{pybamm.exp}, \texttt{pybamm.tanh}, \texttt{pybamm.sqrt}, etc.
  \item Do not use \texttt{pybamm.pi}; if you need pi, use \texttt{3.14} instead.
  \item Try different algorithms; do not use the same or highly similar algorithm multiple times.
  \item The function must be named \texttt{current\_function} and must not take any parameters.
  \item The function must return a PyBaMM expression that represents the charging current as a function of time.
  \item Make sure every Python code block is enclosed by matching triple backticks and starts with \texttt{python}.
\end{enumerate}

Follow the format below when generating the algorithm:

\begin{verbatim}
# Charging Function
```python
def current_function():
    t = pybamm.t
    # Your algorithm here
    return i  # where i is the charging current expression
```    
\end{verbatim}
\end{tcolorbox}

\subsection{P2O prompt}
\label{app:p2o_prompt}

\subsubsection*{Prompt for P2O Initialization Task (Case1)}
\begin{tcolorbox}[promptbox]
Generate 10 reasonable trainable (has parameters to learn) neural network structures using PyTorch with the following constraints:

\begin{enumerate}
    \item \textbf{Framework:} \\
    Use PyTorch. Import \texttt{torch}, \texttt{torch.nn}, and \texttt{torch.nn.functional}.
    \item \textbf{Output Scaling:} \\
    Ensure the final output is scaled to the range [0, 10]. Use a sigmoid activation followed by scaling to achieve this.
    \item \textbf{Parameters Constraint:} \\
    Each neural network should have a single input and a single output.
    \item \textbf{Layer Configuration:}
    \begin{itemize}
        \item Number of layers should range between 0 and 3.
        \item Neurons in each layer should be between 1 and 4.
        \item Consider use different activation functions, normalization layers.
        \item Do not use dropout layers.
    \end{itemize}
    \item \textbf{Network Naming:} \\
    Each generated class should be named \texttt{NeuralNetwork}. When creating instances of these classes, the name \texttt{NeuralNetwork} should always be used.
\end{enumerate}

Below is a template for each network:

\paragraph{Neural Network 1}
\begin{verbatim}
import torch
import torch.nn as nn
import torch.nn.functional as F

class NeuralNetwork(nn.Module):
    def __init__(self):
        super(NeuralNetwork, self).__init__()
        ...
        
    def forward(self, input):
        ...
        output = torch.sigmoid(x) * 10  # Scale output to range [0, 10]
        return output
\end{verbatim}

Repeat this pattern for 10 different networks.
\end{tcolorbox}

\subsubsection*{Prompt for P2O Evolution Task (Case1)}

\begin{tcolorbox}[promptbox]
Based on the two given neural networks, which represent the shape of a charging protocol, make a small and focused adjustment to generate a new neural network. Follow these specific instructions:

\begin{enumerate}
    \item \textbf{Framework}: Use PyTorch. Import \texttt{torch}, \texttt{torch.nn}, and \texttt{torch.nn.functional}.
    
    \item \textbf{Output Scaling}: Ensure the final output is scaled to the range of 0 to 10.
    
    \item \textbf{Parameters Constraint}: The new neural network should have a single input and a single output.
    
    \item \textbf{Network Structure \& Trick}: Try different number of layers and neurons. Incorporate various activation functions, residual connections, and different normalization layers to enhance performance. IMPORTANT: But don't use dropout.

    \item \textbf{Parameter Number}: The parameters of the generated network should be less than 35.
    
    \item \textbf{Structure}: Provide your answer in the following format:
\end{enumerate}

\begin{lstlisting}[language=Python]
import torch
import torch.nn as nn
import torch.nn.functional as F

class NeuralNetwork(nn.Module):
    def __init__(self):
        super(NeuralNetwork, self).__init__()
        ...

    def forward(self, input):
        ...
        output = torch.sigmoid(x) * 10  # Scale output to range [0, 10]
        return output
\end{lstlisting}

\end{tcolorbox}

\subsubsection*{Prompt for P2O Initialization Task (case3)}

\begin{tcolorbox}[promptbox]
You are tasked with generating 10 PyTorch neural networks and their symbolic counterparts in PyBaMM, including RNN, MLP, LSTM, etc. Follow the strict design constraints and code template provided below:

\begin{enumerate}
    \item \textbf{Framework:}
    \begin{itemize}
        \item You must import:
        \begin{verbatim}
import torch
import torch.nn as nn
import torch.nn.functional as F
import pybamm
import numpy as np
        \end{verbatim}
    \end{itemize}

    \item \textbf{Output Scaling:}
    \begin{itemize}
        \item Ensure the final output is scaled to the range [-8, -3].
        \item Use a sigmoid activation followed by scaling to achieve this.
    \end{itemize}

    \item \textbf{Parameters Constraint:}
    \begin{itemize}
        \item Each neural network should have three inputs and a single output.
    \end{itemize}

    \item \textbf{Layer Configuration:}
    \begin{itemize}
        \item Consider different architectures: MLP, RNN, LSTM, etc.
        \item Consider different activation functions.
        \item Do not use dropout layers.
        \item The total number of parameters of each generated network should be less than 35.
    \end{itemize}

    \item \textbf{Network Naming:}
    \begin{itemize}
        \item Each generated PyTorch class should be named \texttt{NeuralNetwork}.
        \item The PyBaMM symbolic version should be a standalone function named \texttt{nn\_in\_pybamm}.
    \end{itemize}
\end{enumerate}

Below is an example and template for each network:

\textbf{Neural Network 1}
\begin{verbatim}
import torch
import torch.nn as nn
import torch.nn.functional as F
import pybamm
import numpy as np

class NeuralNetwork(nn.Module):
    def __init__(self):
        super().__init__()
        self.fc1 = nn.Linear(3, 3)
        self.ln1 = nn.LayerNorm(3, elementwise_affine=False)
        self.act = nn.LeakyReLU(0.01)
        self.fc2 = nn.Linear(3, 3)
        self.fc3 = nn.Linear(3, 1)

    def forward(self, x: torch.Tensor) -> torch.Tensor:
        h = self.fc1(x)
        h = self.ln1(h)
        h = self.act(h)
        h_res = self.fc2(h) + h
        h = self.act(h_res)
        out = -torch.sigmoid(self.fc3(h)) * 5 - 3
        return out

def nn_in_pybamm(X, Ws, bs):
    W1, W2, W3 = [w.tolist() for w in Ws]
    b1, b2, b3 = [b.tolist() for b in bs]
    hidden = len(b1)
    assert hidden == 3
    # --- layer 1: affine ---
    lin1 = []
    for i in range(hidden):
        x = pybamm.Scalar(b1[i])
        for j in range(3):
            x += pybamm.Scalar(W1[i][j]) * X[j]
        lin1.append(x)

    # --- LayerNorm (no affine) ---
    mean1 = sum(lin1) / hidden
    var1 = sum((x - mean1) ** 2 for x in lin1) / hidden
    eps = 1e-5  # small constant to avoid division by zero
    inv_std1 = 1 / pybamm.sqrt(var1 + pybamm.Scalar(eps))
    ln1 = [(x - mean1) * inv_std1 for x in lin1]

    # --- activation ---
    h1 = [pybamm.maximum(x, 0.01 * x) for x in ln1]

    # --- layer 2: affine + residual + activation ---
    lin2 = []
    for i in range(hidden):
        x = pybamm.Scalar(b2[i])
        for j in range(hidden):
            x += pybamm.Scalar(W2[i][j]) * h1[j]
        lin2.append(x)
    h2 = [
        pybamm.maximum(lin2[i] + h1[i], 0.01 * (lin2[i] + h1[i]))
        for i in range(hidden)
    ]

    # --- output layer + sigmoid + scaling ---
    lin3 = pybamm.Scalar(b3[0])
    for j in range(hidden):
        lin3 += pybamm.Scalar(W3[0][j]) * h2[j]
    y = 1 / (1 + pybamm.exp(-lin3))
    I_scale = 5  # Scale output to current range
    return -pybamm.Scalar(I_scale) * y - 3
\end{verbatim}

\textbf{Neural Network 2}
\begin{verbatim}
import torch
import torch.nn as nn
import torch.nn.functional as F
import pybamm
import numpy as np

class NeuralNetwork(nn.Module):
    def __init__(self):
        super(NeuralNetwork, self).__init__()
        ...
        
    def forward(self, input):
        ...
        output = -torch.sigmoid(x) * 5 - 3  
        # Scale output to range [-8, -3]
        return output

def nn_in_pybamm(X, Ws, bs):
    ...
    return -pybamm.Scalar(I_scale) * y - 3
\end{verbatim}

Repeat this pattern for 10 different networks.
\end{tcolorbox}

\subsubsection*{Prompt for P2O Evolution Task (Case3)}

\begin{tcolorbox}[promptbox]
You are tasked with generating a PyTorch neural network and its symbolic counterpart in PyBaMM, based on two given example networks. Follow the strict design constraints and code template provided below:

\begin{enumerate}
    \item \textbf{Framework:}
    \begin{itemize}
        \item You must import:
        \begin{verbatim}
import torch
import torch.nn as nn
import torch.nn.functional as F
import pybamm
import numpy as np
        \end{verbatim}
    \end{itemize}

    \item \textbf{Output Scaling:}
    \begin{itemize}
        \item Ensure the final output is scaled to the range [-8, -3].
        \item Use a sigmoid activation followed by scaling to achieve this.
    \end{itemize}

    \item \textbf{Parameters Constraint:}
    \begin{itemize}
        \item Each neural network should accept three input features per time step and produce a single scalar output.
    \end{itemize}

    \item \textbf{Layer Configuration:}
    \begin{itemize}
        \item Consider different architectures (e.g., MLP, RNN, LSTM, etc.).
        \item Do not use dropout layers.
        \item Vary activation functions (e.g., ReLU, LeakyReLU, ELU, Tanh, etc.).
        \item You may include residual (skip) connections.
        \item The total number of parameters of each generated network should be less than 35.
    \end{itemize}

    \item \textbf{Naming:}
    \begin{itemize}
        \item The PyTorch model class must be named \texttt{NeuralNetwork}.
        \item The PyBaMM symbolic function must be named \texttt{nn\_in\_pybamm}.
    \end{itemize}

\end{enumerate}

Below is the template to follow for each generated network:

\begin{verbatim}
import torch
import torch.nn as nn
import torch.nn.functional as F
import pybamm
import numpy as np

class NeuralNetwork(nn.Module):
    def __init__(self):
        super().__init__()
        # define layers, layer norms, activations here

    def forward(self, x: torch.Tensor) -> torch.Tensor:
        # forward through layers, activations, residuals here
        out = torch.sigmoid(self.final_layer(x)) * 5 - 3
        return out

def nn_in_pybamm(X, Ws, bs):
    # Unpack weights and biases
    # Build symbolic computation
    return -pybamm.Scalar(5) * y - 3
\end{verbatim}

\end{tcolorbox}

\section{Case Study 1}
\subsection{ECM model}
\label{appendix:ecm}

The heat generation $Q$ and temperature update $T$ at each time step are calculated using the following ECM-based equations:

\begin{align}
R_0 &= R_{0,\text{ref}} \left(1 + \alpha (T - T_{\text{ref}}) \right) \\
R_p &= R_{p,\text{ref}} \left(1 + \alpha (T - T_{\text{ref}}) \right) \\
Q &= I^2 (R_0 + R_p) \\
T_{\text{next}} &= T + \Delta t \cdot \frac{I^2(R_0 + R_p) - hA(T - T_{\text{ambient}})}{mc}
\end{align}

\noindent
\textbf{Where:}
\begin{itemize}
    \item $I$: current input [A]
    \item $T$: current temperature [K]
    \item $R_0$, $R_p$: ohmic and polarization resistances [\(\Omega\)]
    \item $R_{0,\text{ref}}$, $R_{p,\text{ref}}$: reference resistances at $T_{\text{ref}}$ [\(\Omega\)]
    \item $\alpha$: temperature coefficient [1/K]
    \item $Q$: heat generated [W]
    \item $h$: heat transfer coefficient [W/m$^2\cdot$K]
    \item $A$: surface area for heat exchange [m$^2$]
    \item $T_{\text{ambient}}$: ambient temperature [K]
    \item $m$: thermal mass [kg]
    \item $c$: specific heat capacity [J/kg$\cdot$K]
    \item $\Delta t$: time step duration [s]
\end{itemize}

\subsection{Outer loop evaluation setup}
\label{app:setup}

To ensure fair and efficient comparison among the three architecture search methods, we chose distinct but related search ranges tailored to each algorithm’s strengths and computational characteristics:

\begin{itemize}

    \item \textbf{Bayesian Optimization (BO)}  
    Operates on a fixed, globally modeled search space with higher per-sample overhead.
    \begin{itemize}
      \item Hidden layers: $\{0,1,2,3,4\}$  
      \item Neurons per layer: $\{1,2,3,4,5\}$  
      \item Activation functions: 
        \{\texttt{ReLU}, \texttt{Sigmoid}, \texttt{Tanh}, \texttt{LeakyReLU}, \texttt{ELU}, \texttt{Softplus}, \texttt{Softsign}, \texttt{SiLU}, \texttt{GELU}\}  
      \item Normalization: \{\texttt{none}, \texttt{BatchNorm1d}, \texttt{LayerNorm}\}  
      \item Optimization backend: Optuna (default \texttt{TPESampler}, no pruning).  
    \end{itemize}

  \item \textbf{Evolutionary Algorithm (EA)}  
    Explores architectures progressively via mutation and crossover, benefiting from a narrower initialization that can expand through evolution.
    \begin{itemize}
      \item Initial hidden layers: $\{0,1,2,3\}$  
      \item Initial neurons per layer: $\{1,2,3,4\}$  
      \item Activation functions (initial and mutation): 
        \{\texttt{ReLU}, \texttt{Sigmoid}, \texttt{Tanh}, \texttt{LeakyReLU}, \texttt{ELU}, \texttt{Softplus}, \texttt{Softsign}, \texttt{SiLU}, \texttt{GELU}\}  
      \item Normalization (initial and mutation): \{\texttt{none}, \texttt{batchnorm}, \texttt{layernorm}\}  
      \item Mutation probability: $p_{\mathrm{mut}} = 0.1$  
      \item Mutation operators:
        \begin{itemize}
          \item Width perturbation: $\pm1$ neuron, clipped to $[1,10]$  
          \item Activation replacement among the nine choices  
          \item Normalization add/remove/change  
        \end{itemize}
      \item Crossover: single-point on layer sequence when both parents have $\ge2$ layers; otherwise local width perturbation  
      \item Selection: tournament size = 2  
    \end{itemize}

  \item \textbf{Random Sampling}  
    Serves as a non-adaptive baseline, drawing architectures independently each generation from the same initial space as EA for direct comparison of adaptive benefits.
    \begin{itemize}
      \item Hidden layers: $\{0,1,2,3\}$  
      \item Neurons per layer: $\{1,2,3,4\}$  
      \item Activation functions: 
        \{\texttt{ReLU}, \texttt{Sigmoid}, \texttt{Tanh}, \texttt{LeakyReLU}, \texttt{ELU}, \texttt{Softplus}, \texttt{Softsign}, \texttt{SiLU}, \texttt{GELU}\}  
      \item Normalization: 20\% chance per layer to apply either \texttt{batchnorm} or \texttt{layernorm}  
      \item Generation size: 3 new models per generation, over 150 generations  
    \end{itemize}

  \item \textbf{Unified Parameter Cap}  
    To prevent over-parameterization and ensure comparability, all methods enforced a hard limit of 35 trainable parameters. Any architecture exceeding this threshold was assigned a fixed loss of $1\times10^{6}$ during evaluation.
\end{itemize}

\subsection{P2P optimal algorithms}

\begin{tcolorbox}[
  enhanced,
  breakable,
  colback=gray!5,             
  colframe=gray!50!black,    
  title={Python implementation},
  fonttitle=\bfseries,
  coltitle=white,
  arc=2mm
]
\begin{verbatim}
def current_function():
    t = pybamm.t
    # RMS-coherence regulator

    # Smooth progress variables
    q = t / (t + 680)
    s = 0.5 * (1 + pybamm.tanh((t - 320) / 560))
    w = pybamm.sqrt(1 + t / 820)

    # Phase carriers
    phi1 = t / (540 + 130 * q) + 
    0.19 * pybamm.sin(2 * 3.14 * t / (930 + 190 * w))
    phi2 = t / (780 + 210 * w) - 
    0.16 * pybamm.cos(2 * 3.14 * t / (670 + 210 * q))
    phi3 = t / (1180 + 280 * s) + 
    0.12 * pybamm.sin(2 * 3.14 * t / (960 + 260 * s))

    u = pybamm.sin(2 * 3.14 * phi1)
    v = pybamm.cos(2 * 3.14 * phi2)
    z = pybamm.sin(2 * 3.14 * phi3)

    # RMS amplitude and cross-coherence
    A = pybamm.sqrt((u * u + v * v + z * z) / 3)
    C = (u * v + v * z + z * u) / 3

    # Composite coherence metric
    H = 0.55 * A + 0.45 * pybamm.exp(-0.7 * C * C)

    # Ripple budget
    r = 0.017 + 0.011 * q * (0.5 + 0.5 * s)

    # Near-constant magnitude with smooth, bounded ripple
    M = 0.90
    g = M * (1 - r) + M * r / (1 + 0.85 * H + 0.15 * A * A)

    # Negative charging current, |i| <= 10 A
    i = -10 * g
    return i
\end{verbatim}
\end{tcolorbox}

\subsection{Case1 P2O optimal neural network}
\label{app:Case1_P2O_nn}

\begin{figure*}[h]
\centering
    \includegraphics[width=1\linewidth]{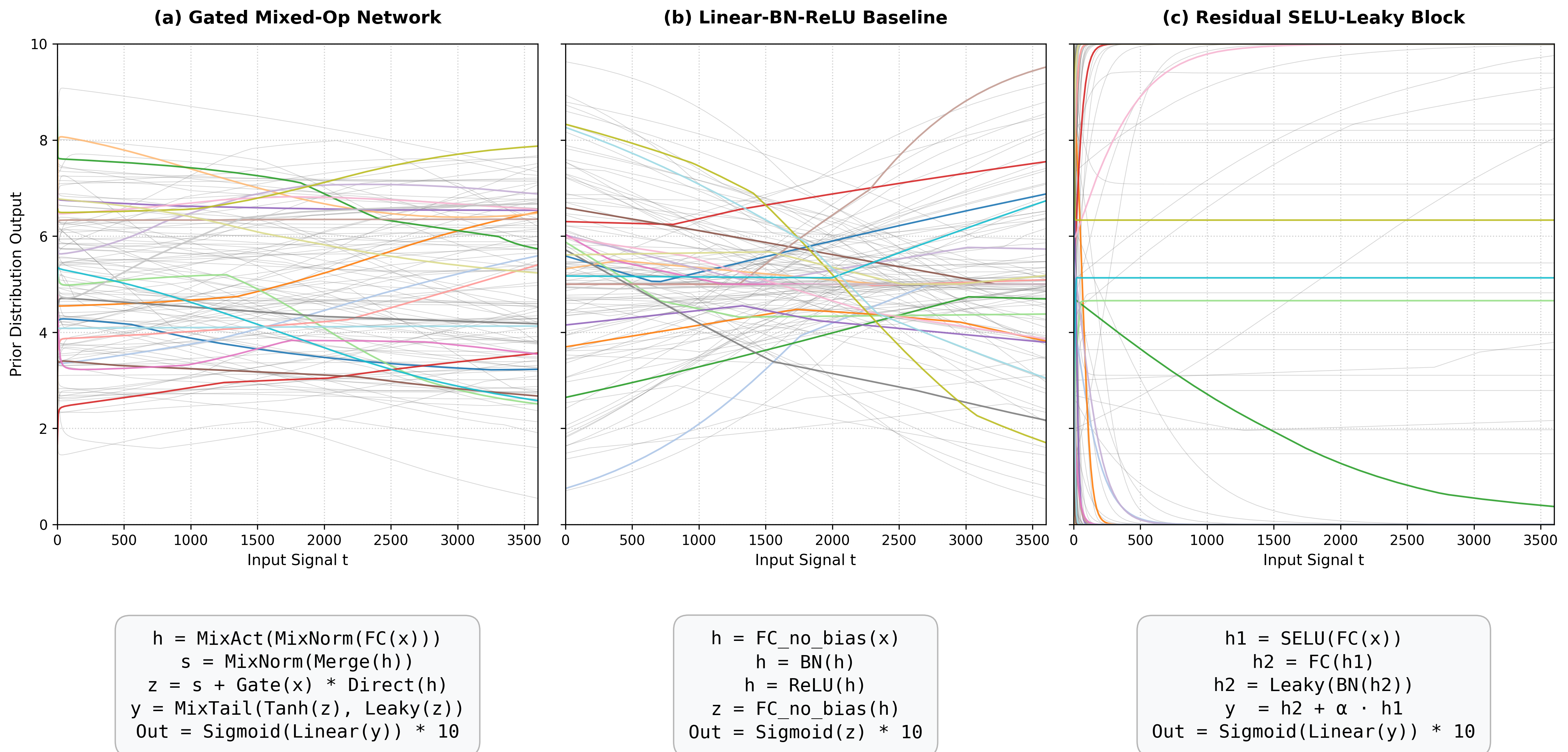}
    \caption{Visualization of architectural functional priors (search spaces).  
The plots illustrate the outputs of randomly initialized networks ($N=120$) over the input signal $t$.  
(a) The Gated Mixed-Op Network exhibits a dense and highly expressive search space.  
(b) The Linear-BN-ReLU Baseline shows a large search space with characteristic piecewise-linear behavior.  
(c) The Residual SELU-Leaky Block demonstrates a more restricted functional prior, visually consistent with the shape of the constant-heating protocol in this battery model.}

    \label{fig:search_space}
\end{figure*}

Fig.\ref{fig:search_space} visualizes the functional priors (search spaces) of the three architectures evolved by the LLM in the constant-heating search. The Gated Mixed-Op Network (Fig. \ref{fig:search_space}a) is structurally more complex, incorporating multiple gating paths and mixed operations, which results in a visibly richer and more intricate search space. The Linear-BN-ReLU Baseline (Fig. \ref{fig:search_space}b) follows a more conventional feed-forward design and therefore exhibits a simpler, although still relatively large, search space. In contrast, the Residual SELU-Leaky Block (Fig. \ref{fig:search_space}c) exhibits a much more constrained and smooth functional prior. Importantly, the optimal protocol in the constant-heating case follows a similarly simple pattern, gradually decreasing and then stabilizing, as shown in Fig. \ref{fig:eval_outer}f. The functional forms favored by this block align well with the structure of the target solution, which is consistent with the strong optimization performance observed in this setting.

\begin{tcolorbox}[title=Neural Network Structure 1, colback=gray!5, colframe=gray!50!black, listing only, listing options={language=Python, breaklines=true}, breakable]

\begin{verbatim}

class NeuralNetwork(nn.Module):
    def __init__(self):
        super(NeuralNetwork, self).__init__()
        # Early stack (from both seeds)
        self.fc1 = nn.Linear(1, 4)
        self.bn1 = nn.BatchNorm1d(4)  # Seed1-style affine BN
        self.ln1 = nn.LayerNorm(4, elementwise_affine=False)

        # Blending scalars: decoupled norms (Seed2) + act + tail
        self.alpha_act = nn.Parameter(torch.tensor(0.5))
        self.alpha_norm1 = nn.Parameter(torch.tensor(0.5))  
        self.alpha_norm2 = nn.Parameter(torch.tensor(0.5))  
        self.alpha_tail = nn.Parameter(torch.tensor(0.5))  

        # Scalar projection (Seed2-style BN2 affine=False)
        self.merge = nn.Linear(4, 1)
        self.bn2 = nn.BatchNorm1d(1, affine=False)
        self.ln2 = nn.LayerNorm(1, elementwise_affine=False)

        # Gating and direct path (common)
        self.gate = nn.Linear(1, 1)
        self.direct = nn.Linear(4, 1)

        # Nonlinearity and final head
        self.leaky = nn.LeakyReLU(0.2)
        self.out = nn.Linear(1, 1)

    def forward(self, input):
        w_act = torch.sigmoid(self.alpha_act)
        w_n1 = torch.sigmoid(self.alpha_norm1)
        w_n2 = torch.sigmoid(self.alpha_norm2)
        w_tail = torch.sigmoid(self.alpha_tail)

        # Early projection + blended normalization + blended activation
        x = self.fc1(input)
        x = (1.0 - w_n1) * self.bn1(x) + w_n1 * self.ln1(x)
        x = (1.0 - w_act) * F.relu(x) + w_act * F.silu(x)

        # Scalar path with separate BN/LN blend
        s = self.merge(x)
        s = (1.0 - w_n2) * self.bn2(s) + w_n2 * self.ln2(s)

        # Input-conditioned gate and direct injection
        z = self.direct(x)
        g = torch.sigmoid(self.gate(input))
        s_tilde = s + g * z

        # Combine Seed1's dual-tail scheme with Seed2's tail mixer
        w_mix_seed1 = 0.5 * (g + w_act)
        y1 = (1.0 - w_mix_seed1) * torch.tanh(s_tilde) 
        + w_mix_seed1 * self.leaky(s_tilde)
        y2 = (1.0 - g) * torch.tanh(s_tilde) + g * self.leaky(s_tilde)

        # Crossfade between tails using both norm2 and tail mixers
        w_cf = 0.5 * (w_tail + w_n2)
        y = (1.0 - w_cf) * y1 + w_cf * y2

        # Final mapping and scaling to [0, 10]
        x = self.out(y)
        output = torch.sigmoid(x) * 10
        return output


\end{verbatim}
\end{tcolorbox}   

\begin{tcolorbox}[title=Neural Network Structure 2, colback=gray!5, colframe=gray!50!black, listing only, listing options={language=Python}]

\begin{verbatim}
class NeuralNetwork(nn.Module):
    def __init__(self):
        super(NeuralNetwork, self).__init__()
        self.fc1 = nn.Linear(1, 3, bias=False)
        self.bn1 = nn.BatchNorm1d(3)
        self.fc2 = nn.Linear(3, 1, bias=False)
        
    def forward(self, input):
        x = self.fc1(input)
        x = self.bn1(x)  # Batch normalization
        x = torch.relu(x)  # Using ReLU activation function
        x = self.fc2(x)
        output = torch.sigmoid(x) * 10  # Scale output to range [0, 10]
        return output
\end{verbatim}
\end{tcolorbox}      

\begin{tcolorbox}[title=Neural Network Structure 3, colback=gray!5, colframe=gray!50!black, listing only, listing options={language=Python, breaklines=true}， breakable]

\begin{verbatim}
class NeuralNetwork(nn.Module):
    def __init__(self):
        super(NeuralNetwork, self).__init__()
        self.fc1 = nn.Linear(1, 3)
        self.fc2 = nn.Linear(3, 3)
        self.bn2 = nn.BatchNorm1d(3)
        self.out = nn.Linear(3, 1)
        self.res_scale = nn.Parameter(torch.tensor(1.0))  

    def forward(self, input):
        h1 = F.selu(self.fc1(input))
        h2 = self.fc2(h1)
        h2 = self.bn2(h2)
        h2 = F.leaky_relu(h2, negative_slope=0.1)
        x = h2 + self.res_scale * h1
        x = self.out(x)
        output = torch.sigmoid(x) * 10 
        return output

\end{verbatim}
\end{tcolorbox}

\section{Case study 2}

\subsection{P2O optimal neural network (Case 2)}
\label{app:case2_nn}
\begin{tcolorbox}[title=Neural Network Structure, colback=gray!5, colframe=gray!50!black, listing only, listing options={language=Python}]

\begin{verbatim}
import torch
import torch.nn as nn
import torch.nn.functional as F


class NeuralNetwork(nn.Module):
    def __init__(self):
        super(NeuralNetwork, self).__init__()
        self.fc1 = nn.Linear(1, 3, bias=False)
        self.bn1 = nn.BatchNorm1d(3)
        self.fc2 = nn.Linear(3, 1, bias=False)
        
    def forward(self, input):
        x = self.fc1(input)
        x = self.bn1(x)  # Batch normalization
        x = torch.relu(x)  # Using ReLU activation function
        x = self.fc2(x)
        output = torch.sigmoid(x) * 10  # Scale output to range [0, 10]
        return output
\end{verbatim}
\end{tcolorbox}      

\subsection{P2P optimal algorithm (Case 2)}
\label{app:case2_p2o_alg}

\begin{tcolorbox}[
  enhanced,
  breakable,
  colback=gray!5,              
  colframe=gray!50!black,     
  title={Python implementation},
  fonttitle=\bfseries,
  coltitle=white,
  arc=2mm
]
\begin{verbatim}
def current_function():
    t = pybamm.t
    Q = 0.4  # target heat (watts)
    R0 = 0.02  # base ohmic resistance (ohms)

    # Smooth bases
    s1 = pybamm.sin(2 * 3.14 * t / 2600 + 0.3)
    c1 = pybamm.cos(2 * 3.14 * t / 3900 - 0.6)
    s2 = pybamm.sin(2 * 3.14 * t / 5200 + 1.0)
    c2 = pybamm.cos(2 * 3.14 * t / 6700 + 0.8)
    s3 = pybamm.sin(2 * 3.14 * t / 8100 - 1.1)
    c3 = pybamm.cos(2 * 3.14 * t / 9500 + 0.2)

    # Nonlinear modulators
    u1 = pybamm.tanh(0.7 * s1 + 0.4 * c2)
    u2 = pybamm.tanh(0.6 * s2 - 0.5 * c1)
    u3 = pybamm.tanh(0.5 * s3 + 0.6 * c3)
    u4 = pybamm.tanh(0.8 * s1 - 0.3 * c3)

    # Positive, time-varying resistances (diverse constructions)
    R1 = R0 * pybamm.exp(0.22 * s1 + 0.10 * c1 + 0.06 * s2 * c2)
    R2 = R0 * (1.05 + 0.25 * s2 + 0.15 * c2 + 0.10 * u1) ** 2
    R3 = R0 * pybamm.exp(0.18 * c3 - 0.08 * s1 + 0.05 * u2)
    R4 = R0 * (0.95 + 0.20 * c1 - 0.10 * s3 + 0.20 * u3) ** 2
    R5 = R0 * pybamm.exp(0.14 * u4 + 0.07 * s2 - 0.04 * c2)
    R6 = R0 * (1.00 + 0.18 * s1 * c1 + 0.12 * s2 * s3) ** 2

    # Smooth softmax weights over six channels
    gamma = 1.0 + 0.3 * pybamm.sin(2 * 3.14 * t / 10400)
    v1 = pybamm.exp(gamma * s1)
    v2 = pybamm.exp(gamma * c1)
    v3 = pybamm.exp(gamma * s2)
    v4 = pybamm.exp(gamma * c2)
    v5 = pybamm.exp(gamma * s3)
    v6 = pybamm.exp(gamma * c3)
    sumv = v1 + v2 + v3 + v4 + v5 + v6
    w1 = v1 / sumv
    w2 = v2 / sumv
    w3 = v3 / sumv
    w4 = v4 / sumv
    w5 = v5 / sumv
    w6 = v6 / sumv

    # Tri-mean over arithmetic, geometric, and harmonic means (A-G-H)
    R_arith = w1 * R1 + w2 * R2 + w3 * R3 + w4 * R4 + w5 * R5 + w6 * R6
    R_harm = 
    1 / (w1 / R1 + w2 / R2 + w3 / R3 + w4 / R4 + w5 / R5 + w6 / R6)
    R_geom = pybamm.exp(
        w1 * pybamm.log(R1)
        + w2 * pybamm.log(R2)
        + w3 * pybamm.log(R3)
        + w4 * pybamm.log(R4)
        + w5 * pybamm.log(R5)
        + w6 * pybamm.log(R6)
    )

    delta = 0.7
    x1 = pybamm.exp(delta * pybamm.sin(2 * 3.14 * t / 7200 + 0.4))
    x2 = pybamm.exp(delta * pybamm.cos(2 * 3.14 * t / 7800 - 0.9))
    x3 = pybamm.exp(delta * pybamm.sin(2 * 3.14 * t / 8400 + 1.1))
    sumx = x1 + x2 + x3
    a = x1 / sumx
    b = x2 / sumx
    c = x3 / sumx

    R_AHG = (R_arith ** a) * (R_geom ** b) * (R_harm ** c)

    # Conductance-side power-harmonic mean, mapped back to resistance
    G1 = 1 / R1
    G2 = 1 / R2
    G3 = 1 / R3
    G4 = 1 / R4
    G5 = 1 / R5
    G6 = 1 / R6
    r = 0.6 + 0.4 * pybamm.sin(2 * 3.14 * t / 8800 - 0.2)  
    denom = (
        w1 / (G1 ** r)
        + w2 / (G2 ** r)
        + w3 / (G3 ** r)
        + w4 / (G4 ** r)
        + w5 / (G5 ** r)
        + w6 / (G6 ** r)
    )
    G_ph = (1 / denom) ** (1 / r)
    R_condpath = 1 / G_ph

    # Harmonic blending between A-G-H path and conductance path
    alpha = 0.5 + 0.5 * pybamm.sin(2 * 3.14 * t / 7600 + 0.3)
    R_tilde = 1 / (alpha / R_AHG + (1 - alpha) / R_condpath)

    # Smooth envelope to diversify dynamics
    env = pybamm.exp(0.12 * 
    pybamm.tanh(0.5 * s1 - 0.4 * c3 + 0.3 * s2 * c1))
    R_t = R_tilde * env

    # Current targeting constant heat:
    # i^2 * R_t \approx Q, saturated to [-5, 0] A
    i_mag = pybamm.sqrt(Q / R_t)
    i = -5 * pybamm.tanh(i_mag / 5)
    return i
\end{verbatim}
\end{tcolorbox}

\section{Case study 3}

\subsection{Degradation submodels}
\label{app:case3_model}
The following degradation submodels were enabled in PyBaMM for the DFN simulations:
\begin{itemize}
    \item \textbf{Particle}: Fickian diffusion
    \item \textbf{Particle mechanics}: swelling and cracking
    \item \textbf{Loss of active material}: stress-driven
    \item \textbf{SEI}: reaction-limited
    \item \textbf{SEI porosity change}: true
    \item \textbf{SEI on cracks}: true
    \item \textbf{Thermal model}: lumped
    \item \textbf{Discharge energy calculation}: enabled
\end{itemize}

\subsection{Accelerated aging parameter modifications}
\label{app:case3_acc}
To accelerate degradation, several parameters were modified relative to the default 
\texttt{Ai2020} values. These are summarized in Table~\ref{tab:acc_params}.

\begin{table}[h]
\centering
\caption{Accelerated aging parameters (relative to \texttt{Ai2020} baseline).}
\label{tab:acc_params}
\begin{tabular}{ll}
\hline
\textbf{Parameter} & \textbf{Modified value} \\
\hline
SEI reaction exchange current density [A m$^{-2}$] & $\times 5$ \\
Negative electrode LAM constant proportional term [s$^{-1}$] & $1.0 \times 10^{-12}$ \\
Positive electrode LAM constant proportional term [s$^{-1}$] & $2.78 \times 10^{-13} \times 10$ \\
Positive electrode cracking rate & $3.9 \times 10^{-20} \times 10$ \\
Negative electrode cracking rate & $3.9 \times 10^{-20} \times 10$ \\
Total heat transfer coefficient [W m$^{-2}$ K$^{-1}$] & $5.0$ \\
Ambient temperature [K] & $308.15$ \\
\hline
\end{tabular}
\end{table}

\subsection{Validation of accelerated aging}
\label{app:no_acc}
To verify that accelerated aging preserves the relative ranking of charging protocols, we repeated simulations for three representative protocols under non-accelerated conditions for 1000 cycles. Results confirmed that protocol rankings were consistent between accelerated and non-accelerated cases (see Fig.\ref{fig:long_run}).

\begin{figure}[h]
\centering
    \includegraphics[width=0.65\linewidth]{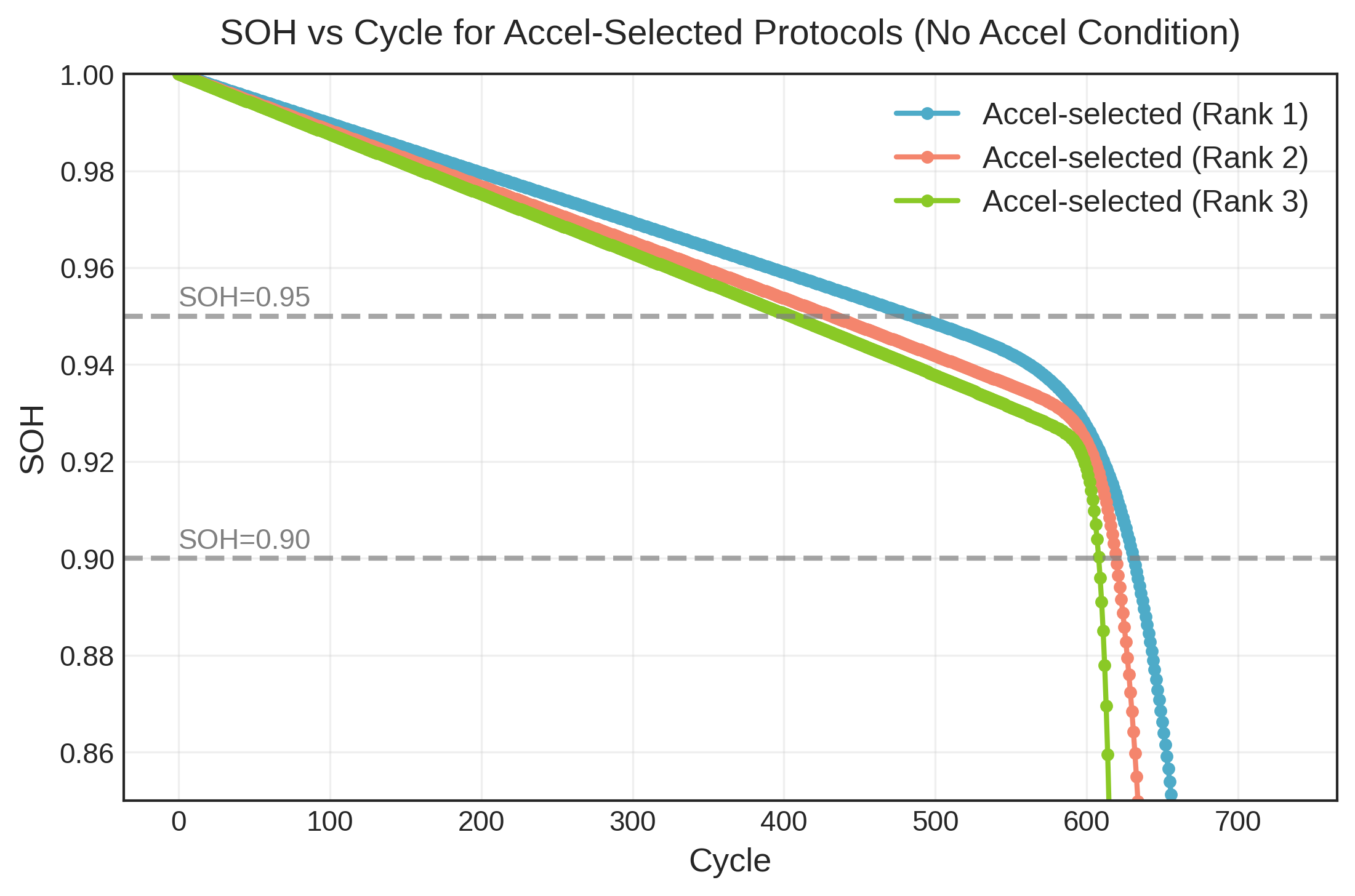}
    \caption{Validation of acceleration factor via long-term non-accelerated simulations.  To enable observable degradation within limited simulation time, aging was artificially accelerated by increasing the SEI exchange current density by a factor of 5.  To verify that this acceleration preserves the relative ranking among charging protocols, three representative protocols selected under accelerated aging were simulated over 1000 cycles under non-accelerated conditions.  The ranking consistency across protocols confirms that the applied acceleration factor preserves protocol differentiation.
    }
    \label{fig:long_run}
\end{figure}

\subsection{Penalty on high voltage}
\label{app:voltage_penalty}
\begin{equation}
\mathcal{P}^{(i)} \;=\; \alpha \int_{0}^{t_{\mathrm{f}}^{(i)}} 
\left[\max\!\left(V(t)-V_{\mathrm{thr}},\,0\right)\right]^n \,\mathrm{d}t,
\end{equation}

where $V_{\mathrm{thr}} = 4.20\,\mathrm{V}$, $n=3$ controls the steepness of the penalty, and $\alpha=0.3$ is a calibrated scaling factor. The penalty term strongly discourages protocols that sustain voltages above $4.20,\mathrm{V}$, thereby biasing the optimization toward voltage-safe strategies. This penalty does not correspond to a directly measured capacity fade; instead, it serves as a surrogate degradation metric introduced to guide the optimization. From a physical perspective, sustained exposure to high voltages accelerates parasitic side reactions such as electrolyte oxidation and transition–metal dissolution, which are known to contribute to long-term capacity loss. Therefore, penalizing the duration and magnitude of voltage overshoot provides a practical proxy for discouraging degradation-prone charging behaviors. At the end of each cycle, the state-of-health (SOH) is updated by combining the measured capacity $Q_{\mathrm{cap}}^{(i)}$ with this penalty term:

\begin{equation}
\mathrm{SOH}^{(i)} \;=\; \frac{Q_{\mathrm{cap}}^{(i)} - \mathcal{P}^{(i)}}{Q_\text{nom}},
\end{equation}

\subsection{P2P optimal algorithm (Case 3)}
\label{app:case3_p2o_alg}

\begin{tcolorbox}[
  enhanced,
  breakable,
  colback=gray!5,              
  colframe=gray!50!black,     
  title={Python implementation},
  fonttitle=\bfseries,
  coltitle=white,
  arc=2mm
]
\begin{verbatim}
def adaptive_current(vars_dict):
    V = vars_dict["Voltage [V]"]              
    T = vars_dict["Volume-averaged cell temperature [K]"]
    SOC = vars_dict["SoC"]

    # Smoothstep shaping in V and SoC (cubic 3x^2 - 2x^3)
    x_v = pybamm.minimum(1, pybamm.maximum(0, 0.5 + (4.08 - V) / 0.35))
    s_v = x_v**2 * (3 - 2 * x_v)
    x_s = pybamm.minimum(1, pybamm.maximum(0, 0.5 + (0.60 - SOC) / 0.50))
    s_s = x_s**2 * (3 - 2 * x_s)

    base = 0.55 * s_v + 0.45 * s_s + 0.50 
    * s_v * s_s - 0.12 * s_v * (1 - s_s)
    base_clamped = pybamm.minimum(1, pybamm.maximum(0, base))
    shape = 1 - pybamm.exp(-2.6 * base_clamped)

    # Temperature modulation: Gaussian around 308.15 K + mild sinusoid
    zT = (T - 308.15) / 17
    t_fac = 0.92 + 0.20 * pybamm.exp(-(zT ** 2)) 
    + 0.06 * pybamm.sin((T - 308.15) / 22)

    # Voltage penalties above 4.20 V and near 4.25 V (softplus hinges)
    v_soft = pybamm.log(1 + pybamm.exp(70 * (V - 4.20))) / 70
    v_pen1 = pybamm.exp(-90 * v_soft) / (1 + 25 * v_soft) + 0.04
    v_cap = pybamm.log(1 + pybamm.exp(140 * (V - 4.25))) / 140
    v_pen = v_pen1 * pybamm.exp(-160 * v_cap)

    # Desired magnitude and clamp
    desired = (3.0 + 6.6 * shape) * t_fac * v_pen
    mag = pybamm.minimum(8, pybamm.maximum(3, desired))
    return -mag
\end{verbatim}
\end{tcolorbox}

\subsection{P2O optimal neural network (Case 3)}
\label{app:case3_p2o_nn}

\begin{tcolorbox}[
  enhanced,
  breakable,
  colback=gray!5,              
  colframe=gray!50!black,     
  title={Python implementation},
  fonttitle=\bfseries,
  coltitle=white,
  arc=2mm
]
\begin{verbatim}
import torch
import torch.nn as nn
import torch.nn.functional as F
import pybamm
import numpy as np


class NeuralNetwork(nn.Module):
    def __init__(self):
        super().__init__()
        # Tiny MLP with a residual skip from input to output
        self.fc1 = nn.Linear(3, 2, bias=True)        # 8 params
        self.ln1 = nn.LayerNorm(2, elementwise_affine=False)  # 0 params
        self.fc_out = nn.Linear(2, 1, bias=True)     # 3 params
        self.skip = nn.Linear(3, 1, bias=False)      # 3 params
        # Total trainable params: 14 (< 35)

    def forward(self, x: torch.Tensor) -> torch.Tensor:
        h = self.fc1(x)
        h = self.ln1(h)
        h = torch.tanh(h)
        y_main = self.fc_out(h)
        y = y_main + self.skip(x)  # residual connection
        out = -torch.sigmoid(y) * 5 - 3  # scale to [-8, -3]
        return out


def nn_in_pybamm(X, Ws, bs):
    """
    Build the symbolic counterpart in PyBaMM.

    Arguments:
    - X: list-like of 3 PyBaMM symbols [x0, x1, x2]
    - Ws: [W1(2x3), W2(1x2), Wskip(1x3)]
    - bs: [b1(2), b2(1)]
    """
    W1, W2, Wskip = [w.tolist() for w in Ws]
    b1, b2 = [b.tolist() for b in bs]

    # z1 = W1 * X + b1  -> shape (2,)
    z1 = []
    for i in range(2):
        zi = pybamm.Scalar(b1[i])
        for j in range(3):
            zi += pybamm.Scalar(W1[i][j]) * X[j]
        z1.append(zi)

    # LayerNorm over 2 features (no affine): 
    # (z - mean) / sqrt(var + eps)
    mean1 = (z1[0] + z1[1]) / 2
    var1 = ((z1[0] - mean1) ** 2 + (z1[1] - mean1) ** 2) / 2
    inv_std1 = 1 / pybamm.sqrt(var1 + pybamm.Scalar(1e-5))
    h1 = [(z1[i] - mean1) * inv_std1 for i in range(2)]

    # Activation: tanh
    a1 = [pybamm.tanh(h1[i]) for i in range(2)]

    # Main head: y_main = W2 * a1 + b2
    y_main = pybamm.Scalar(b2[0])
    for i in range(2):
        y_main += pybamm.Scalar(W2[0][i]) * a1[i]

    # Residual skip: y_skip = Wskip * X (no bias)
    y_skip = pybamm.Scalar(0)
    for j in range(3):
        y_skip += pybamm.Scalar(Wskip[0][j]) * X[j]

    # Combine and scale with sigmoid to [-8, -3]
    y_lin = y_main + y_skip
    y_sig = 1 / (1 + pybamm.exp(-y_lin))
    return -pybamm.Scalar(5) * y_sig - 3

\end{verbatim}
\end{tcolorbox}

\subsection{35 parameter CC optimization}
\label{app:35cc}
\begin{figure}[h]
\centering
    \includegraphics[width=0.65\linewidth]{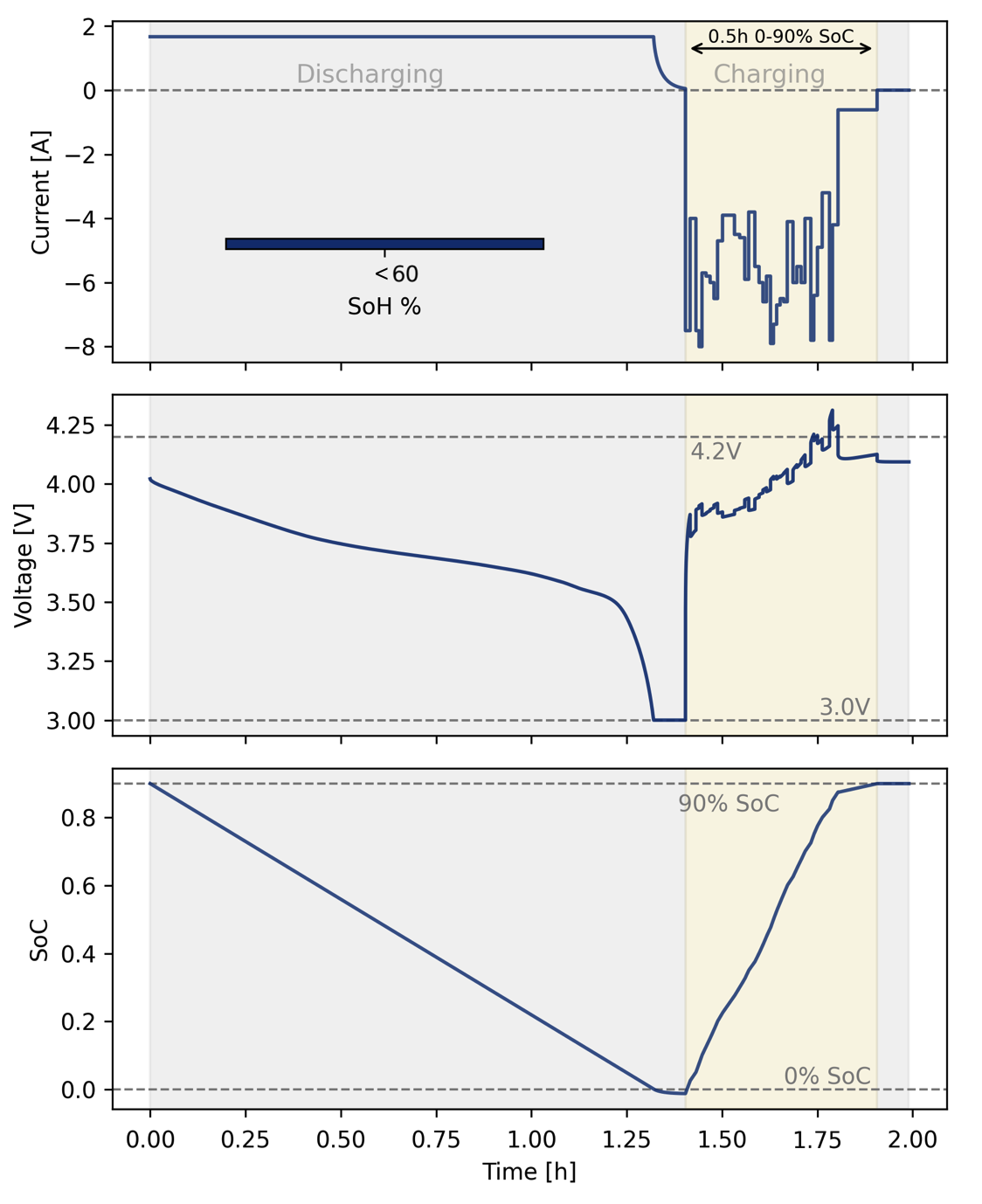}
    \caption{Charging simulation results using the modified multi-segment CC method with 35 segments. Despite the increased number of parameters, the results exhibit pronounced sawtooth patterns—highlighting the inherent instability of the approach when scaled. 
    }
    \label{fig:long_run}
\end{figure}

\clearpage


\end{document}